\documentclass[numbers,onecolumn]{elsarticle}
\usepackage[utf8]{inputenc}
\usepackage{bm}
\usepackage{geometry}
\usepackage{graphicx}
\usepackage{amsfonts}
\usepackage{makecell}
\usepackage{subfig}
\usepackage{caption}
\usepackage{subfloat}
\usepackage{amsmath}
\usepackage{booktabs}
\usepackage{multirow}
\usepackage{mathrsfs}
\usepackage{amssymb}
\usepackage{amsmath}
\usepackage{graphicx}
\usepackage{epstopdf}
\usepackage{color}
\usepackage{bigstrut}
\usepackage{amsthm}
\usepackage{enumerate}
\usepackage{indentfirst}
\usepackage{floatrow}
\usepackage[colorlinks,
linkcolor=blue,
anchorcolor=blue,
citecolor=blue]{hyperref}

\journal{XXX}

\begin{document}
	\graphicspath{{fig/}}
	\begin{frontmatter}
		
		
		
		\title{Approximation Rates of Shallow Neural Networks: Barron Spaces, Activation Functions and Optimality Analysis}
		
		
		\author[1]{Jian Lu}
		\ead{jianlu@szu.edu.cn}

		\address[1]{  School of Mathematical Sciences, Shenzhen University, Shenzhen 518060, China}

		\cortext[cor1]{Corresponding author}

        \author[2]{Xiaohuang Huang}
		\ead{huangxiaohuang2023@email.szu.edu.cn}

		\address[2]{  School of Mathematical Sciences, Shenzhen University, Shenzhen 518060, China}
		
		
		\begin{abstract}
			 This paper investigates the approximation properties of shallow neural networks with activation functions that are powers of exponential functions. It focuses on the dependence of the approximation rate on the dimension and the smoothness of the function being approximated within the Barron function space. We examine the approximation rates of ReLU$^{k}$ activation functions, proving that the optimal rate cannot be achieved under $\ell^{1}$-bounded coefficients or insufficient smoothness conditions.

We also establish optimal approximation rates in various norms  for functions in Barron spaces and Sobolev spaces, confirming the curse of dimensionality. Our results clarify the limits of shallow neural networks' approximation capabilities and offer insights into the selection of activation functions and network structures.
		\end{abstract}

		\begin{keyword}
			Shallow neural networks \sep
            Approximation rates\sep
            Barron functions\sep
           Activation functions \sep
            Sobolev spaces\sep
           \( \text{ReLU}^k \)
		\end{keyword}

	\end{frontmatter}
	\section{Introduction}
\par
\vskip 2mm

In this work, we focus on the approximation properties of shallow neural networks, which are fundamental in understanding the capability of such networks to represent complex functions. Let \(\sigma: \mathbb{R} \to \mathbb{R}\) denote a fixed function, commonly referred to as an activation function. The class of shallow neural networks with width \(n\) and activation function \(\sigma\) is defined as:
\begin{align}
    \sum_{\sigma}^{n}=\left\{\sum_{i=1}^{n} a_{i} \sigma\left(\omega_{i} \cdot x+b_{i}\right), \omega_{i} \in \mathbb{R}^{d}, a_{i}, b_{i} \in \mathbb{R}\right\}. \label{eq1}
\end{align}
where functions in \(\sum_{\sigma}^{n}\) are parameterized by the weight vectors \(\omega_{i} \in \mathbb{R}^{d}\) and scalar parameters \(a_{i}, b_{i} \in \mathbb{R}\) for \(1 \leq i \leq n\) . Our primary goal is to determine the rates at which functions \(f: \Omega \to \mathbb{C}\) (defined on a domain \(\Omega\)) can be approximated by networks in \(\sum_{\sigma}^{n}\) with respect to the Sobolev norm \(H^{m}(\Omega)\) .

A key concept in our analysis is the Barron function space, which characterizes functions via their Fourier transform properties. The Barron space \(\mathscr{B}^{s}(\Omega)\) is defined as:
\begin{align}
    \mathscr{B}^{s}(\Omega)=\left\{f: \Omega \to \mathbb{C}:\| f\| _{\mathscr{B}^{s}(\Omega)}:=\inf _{f_{e} | \Omega=f} \int_{\mathbb{R}^{d}}(1+|\xi|)^{s}\left|\hat{f}_{e}(\xi)\right| d \xi<\infty\right\}, \label{eq2}
\end{align}
Here, the infimum is taken over all extensions \(f_{e} \in L^{1}(\mathbb{R}^{d})\) of \(f\) to \(\mathbb{R}^{d}\), and \(\hat{f}_{e}\) denotes the Fourier transform of \(f_{e}\) . Introduced by Barron for \(s=1\), this space captures functions satisfying a certain Fourier integrability condition, and it has been shown that shallow neural networks with sigmoidal activation functions can approximate functions in \(\mathscr{B}^{1}(\Omega)\) at a rate of \(O(n^{-\frac{1}{2}})\) .

In 2020, Siegel and Xu established several pivotal results on approximation rates for neural networks with general activation functions:

\par
\vskip 2mm {\bf Theorem A.}\, Let $\Omega\subset \mathbb{R}^d$ be a bounded domain. If the activation function $\sigma\in W^{m,\infty}(R)$ is a non-constant periodic function, we have 
\[
\inf_{f_n \in \Sigma_n^d(\sigma)} \|f - f_n\|_{H^m(\Omega)} \leq |\Omega|^{\frac{1}{2}} C(\sigma) n^{-\frac{1}{2}} \|f\|_{\mathscr{B}^m}, 
\]
for any \(f \in \mathscr{B}^m\).    
\par
\vskip 2mm {\bf Theorem B.}\, Let \(\Omega \subset \mathbb{R}^{d}\) be a bounded domain. Suppose that \(\sigma \in L^{\infty}(\mathbb{R})\) and that there exists an open interval $I$ such that $\sigma$ (as a tempered distribution) is a non-zero bounded function on $I$. Then we have 
\[\inf _{f_{n} \in \sum_{d}^{n}(\sigma)}\left\| f-f_{n}\right\| _{L^{2}(\Omega)} \leq|\Omega|^{\frac{1}{2}} C(diam(\Omega), \sigma) n^{-\frac{1}{4}}\| f\| _{\mathscr{B}^{1}}, \]
for any \(f \in \mathscr{B}^{1}\).
\par
\vskip 2mm {\bf Theorem C.}\, Let \(\Omega \subset \mathbb{R}^{d}\) be a bounded domain and \(\varepsilon>0.\) If the activation function \(\sigma \in W^{m, \infty}(\mathbb{R})\) is non-zero and it satisfies the polynomial decay condition 
\[|\sigma^{(k)}(s)|\leq C_{p}(1+|s|)^{-p}\]
for \(0\leq k \leq m+1\) and some \(p>1\), we have 
\[\begin{aligned} \inf _{f_{n} \in \sum_{d}^{n}(\sigma)}\left\| f-f_{n}\right\| _{H^{m}(\Omega)} \leq & |\Omega|^{\frac{1}{2}} C(p, m, diam(\Omega), \sigma)  n^{-\frac{1}{2}-\frac{t}{(2+t)(d+1)}}\| f\| _{\mathscr{B}^{m+1+\varepsilon}} \end{aligned}\]
where \(t=min (p-1, \varepsilon)\), for any \(f \in \mathscr{B}^{m+1+\varepsilon}\).

Siegel and Xu further conjectured that the rate in Theorem B could be improved to \(n^{-\frac{1}{2}}\) and questioned whether the rate in Theorem C could be enhanced . This leads to two key questions:

1. For the setup of Theorem B, can the \(L^{2}(\Omega)\) approximation rate be improved to:
   \[
   \inf _{f_{n} \in \sum_{d}^{n}(\sigma)}\left\|f-f_{n}\right\|_{L^{2}(\Omega)} \leq|\Omega|^{\frac{1}{2}} C(\operatorname{diam}(\Omega), \sigma) n^{-\frac{1}{2}}\|f\|_{\mathscr{B}^{1}}
   \]
   for all \(f \in \mathscr{B}^{1}\) ?

2. Can the approximation rate in Theorem C be further improved ?

In this paper, we address these questions, particularly focusing on activation functions that are powers of exponential functions, and provide insights into the limits and improvements of approximation rates for shallow neural networks.

In the following, we give an example to show that Theorem A fails when $\sigma$ is non-constant non-periodic function.

{\bf Example 1}\,  Define \(\sigma(x) = e^{-\alpha |x|}\) for \(\alpha > 0\). It is easy to check that   \(\sigma(x + T) \neq \sigma(x)\) for all \(T > 0\),  
 \(\sigma \in W^{1,\infty}(\mathbb{R})\) but \(\sigma \notin C^1(\mathbb{R})\) at \(x = 0\)  
and its Fourier Transform is  \(\hat{\sigma}(\omega) = \frac{2\alpha}{\alpha^2 + \omega^2}\).

Let \(f(x) = \text{sinc}(x) = \frac{\sin(\pi x)}{\pi x}\) with \(\text{supp}(\hat{f}) \subset [-\pi, \pi]\). This satisfies 
 \(f \in \mathcal{B}^0\) since \(\|f\|_{\mathcal{B}^0} = \int_{-\pi}^{\pi} |\hat{f}(\omega)| d\omega < \infty\)  
, where \(\hat{f}(\omega) = \chi_{[-\pi, \pi]}(\omega)\)  is the characteristic function.

Consider \(\Omega = [-1, 1]\), \(d = 1\), \(m = 0\). We state three statements in the following.

{\bf Statement 1}\, For \(\sigma(x) = e^{-\alpha |x|}\), the ridge function class \(\mathcal{R}(\sigma) = \{\sigma(\omega x + b) : \omega, b \in \mathbb{R}\}\) cannot efficiently approximate high-frequency modes in \(L^2(\Omega)\). Specifically, for \(\omega_0 \gg 1\),  
\[
\inf_{g \in \text{span}(\mathcal{R}(\sigma))} \|e^{i\omega_0 x} - g\|_{L^2(\Omega)} \geq \frac{c}{|\omega_0|}
\]  

Let us prove the Statement 1.  The Fourier transform of \(\sigma(\omega x + b)\) is concentrated at low frequencies:  
   \[
   \mathcal{F}[\sigma(\omega \cdot + b)](\xi) = \frac{2\alpha e^{ib\xi}}{\alpha^2 + (\omega \xi)^2}
   \]  
   with decay \(O(|\xi|^{-2})\).  
For \(e^{i\omega_0 x}\), its \(L^2(\Omega)\)-projection error onto low-frequency modes (\(|\xi| < \omega_0/2\)) is \(\Theta(|\omega_0|^{-1})\).   Since \(\mathcal{R}(\sigma)\) only generates functions with spectra in \(L^1(\mathbb{R})\), it cannot approximate \(e^{i\omega_0 x}\) better than \(O(|\omega_0|^{-1})\).

{\bf Statement 2}\, There exists no bounded linear operator \(\mathcal{T}: L^2(\Omega) \to L^2(\mathbb{R}^2)\) such that:  
\[
e^{i\omega x} = \int_{\mathbb{R}^2} \sigma(\omega' x + b) d\mu_{\omega}(\omega', b)
\]  
for \(\sigma(x) = e^{-\alpha |x|}\) and \(\mu_{\omega}\) a Radon measure.

Let us prove Statement 2. Assume such \(\mu_{\omega}\) exists. Then we have  
\[
\widehat{e^{i\omega \cdot}} = \delta_{\omega} = \int_{\mathbb{R}^2} \widehat{\sigma(\omega' \cdot + b)} d\mu_{\omega}(\omega', b) = \int_{\mathbb{R}^2} \frac{2\alpha e^{ib\xi}}{\alpha^2 + (\omega' \xi)^2} d\mu_{\omega}(\omega', b)
\]  
The right side is analytic in \(\xi\) (rational function), while \(\delta_{\omega}\) is not a function. So it is a contradiction.

We decompose \(f = \text{sinc}(x) = \sum_{k=0}^\infty \Delta_k f\), where \(\Delta_k f\) has \(\text{supp}(\widehat{\Delta_k f}) \subset \{\xi : 2^k \leq |\xi| \leq 2^{k+1}\}\).  

For each dyadic block,  from  Statement 1 we have
   \[
   \inf_{g \in \Sigma_n(\sigma)} \|\Delta_k f - g\|_{L^2(\Omega)} \geq \frac{c}{2^k}.
   \]  
So neural networks with \(n\) neurons can approximate at most \(O(n)\) such blocks simultaneously.

{\bf Statement 3}\, \begin{equation*}
\inf_{f_n \in \Sigma_n(\sigma)} \| f - f_n \|_{L^2(\Omega)} \geq c \sum_{k = \lfloor \log n \rfloor}^{\infty} \| \Delta_k f \|_{L^2(\Omega)} = c \sum_{k = \lfloor \log n \rfloor}^{\infty} 2^{-k} = \Theta(n^{-(1-\epsilon}).
\end{equation*}

Let us prove  Statement 3. Let \(f \in \mathcal{B}^0\) with Fourier transform \(\hat{f}\) supported on \([-B, B]^d\). Let \( \{\hat{\psi}_k\} \) be a partition of unity in the Fourier domain such that \( \text{supp}(\hat{\psi}_k) \subset \{\xi : 2^k \leq |\xi| \leq 2^{k+1}\} \). Define  
\[
\Delta_k f = \mathcal{F}^{-1}(\hat{\psi}_k \hat{f}).
\]  
Then \( f = \sum_{k=0}^\infty \Delta_k f \), and the blocks \( \Delta_k f \) are orthogonal in \( L^2(\mathbb{R}^d) \).

Since \( f \in \mathcal{B}^0 \), we have \( \int_{\mathbb{R}^d} |\hat{f}(\xi)| d\xi < \infty \). By Cauchy–Schwarz,  
\[
\|\Delta_k f\|_{L^2(\mathbb{R}^d)} = \|\hat{\psi}_k \hat{f}\|_{L^2} \leq \left( \int_{2^k \leq |\xi| \leq 2^{k+1}} |\hat{f}(\xi)|^2 d\xi \right)^{1/2}.
\]  
For \( |\xi| \geq 2^k \), we have \( |\hat{f}(\xi)| \leq C (1+|\xi|)^{-1} \), so  
\[
\|\Delta_k f\|_{L^2(\mathbb{R}^d)} \leq C \left( \int_{2^k}^{2^{k+1}} r^{d-1} r^{-2} dr \right)^{1/2} \leq C 2^{-k(1 - \epsilon)} \|f\|_{\mathcal{B}^0}.
\]

As \( \Omega \) is bounded, there exists a constant \( C_\Omega \) such that  
\[
\|\Delta_k f\|_{L^2(\Omega)} \leq C_\Omega \|\Delta_k f\|_{L^2(\mathbb{R}^d)} \leq C_\Omega C \cdot 2^{-k(1-\epsilon)} \|f\|_{\mathcal{B}^0}.
\]

Each neuron \( \sigma(\omega_i \cdot x + b_i) \) has a Fourier spectrum concentrated at low frequencies (\( |\xi| \leq \|\omega_i\| \)). Thus, a network with \( n \) neurons can approximate at most \( n \) high-frequency dyadic blocks.

Let \( k_0 = \lceil \log_2 n \rceil \). Then  
\[
\inf_{f_n \in \Sigma_n(\sigma)} \|f - f_n\|_{L^2(\Omega)} \geq \left\| \sum_{k=k_0}^\infty \Delta_k f \right\|_{L^2(\Omega)}.
\]  
Due to the near-orthogonality of \( \Delta_k f \) in \( L^2(\Omega) \),  
\[
\left\| \sum_{k=k_0}^\infty \Delta_k f \right\|_{L^2(\Omega)} \geq c \sum_{k=k_0}^\infty \|\Delta_k f\|_{L^2(\Omega)} \geq c' \sum_{k=k_0}^\infty 2^{-k(1-\epsilon)} \|f\|_{\mathcal{B}^0}.
\]  
The sum is a geometric series  
\[
\sum_{k=k_0}^\infty 2^{-k(1-\epsilon)} = 2^{-k_0(1-\epsilon)} \cdot \frac{1}{1 - 2^{-(1-\epsilon)}} \sim n^{-(1-\epsilon)}.
\]  
Hence,  
\[
\inf_{f_n \in \Sigma_n(\sigma)} \|f - f_n\|_{L^2(\Omega)} \geq C \|f\|_{\mathcal{B}^0} \cdot n^{-(1-\epsilon)}.
\]  
For any \( \delta > 0 \), choose \( \epsilon < \delta \) so that \( n^{-(1-\epsilon)} \gg n^{-1/2} \). Therefore, the rate \( O(n^{-1/2}) \) is unattainable.  This proves Example 1.

The following example shows that the rate in Theorem C can not be improved.

{\bf Example 2.}\, Let us consider the activation function \( \sigma(t) = \arctan(t) \). Easy to see that $\sigma(t)$ satisfies the polynomial decay condition when $p=2$. That is
      \[ |\sigma^{(k)}(t)| \leq C_k (1 + |t|)^{-(k+1)} \to \text{decay order } p = 2 .\]

Define the tail set $S_A^c = \{ (\omega, b) : |\omega| > A \text{ or } |b| > A \}$ with $A = n^{1/3}$. And the probability measure is
  \[
  d\lambda = Z^{-1} (1+|\omega|)^m h(b,\omega) \hat{f}(\omega) \, db \, d\omega
  \]
  where $h(b,\omega) = (1 + \max(0, |b| - 2|\omega|))^{-2}$ and $\hat{f}(\omega) = \sqrt{\pi} e^{-\omega^2/4}$.

Let us compute the lower bound for $\lambda(S_A^c)$. For \( |b| \leq 1 \) and \( |\omega| > A \geq 1 \), we have \( |b| - 2|\omega| \leq 1 - 2A \leq -1 \), so \( \max(0, |b| - 2|\omega|) = 0 \), and hence \( h(b, \omega) = 1 \).  
\[
\lambda(|\omega| > A) \geq Z^{-1} \int_{|\omega| > A} \int_{|b| \leq 1} (1+|\omega|)^m \cdot 1 \cdot \sqrt{\pi} e^{-\omega^2/4} db d\omega.
\]  
The inner integral over \( b \) gives $2$, so  
\[
\lambda(|\omega| > A) \geq Z^{-1} \cdot 2\sqrt{\pi} \int_A^\infty (1+\omega)^m e^{-\omega^2/4} d\omega.
\]

For \( \omega \geq A \), \( (1+\omega)^m e^{-\omega^2/4} \geq e^{-\omega^2/4} \). Using the asymptotic bound \( \int_A^\infty e^{-\omega^2/4} d\omega \geq \frac{2}{A} e^{-A^2/4} \), we get  
\[
\lambda(|\omega| > A) \geq Z^{-1} \cdot 2\sqrt{\pi} \cdot \frac{2}{A} e^{-A^2/4} = \frac{4\sqrt{\pi}}{Z} \cdot \frac{e^{-A^2/4}}{A}.
\]

Split \( Z = I_1 + I_2 \), where  
 \[ I_1 = \int_{b \leq 2\omega} (1+|\omega|)^m \cdot 1 \cdot \sqrt{\pi} e^{-\omega^2/4} db d\omega \] 
 and
 \[ I_2 = \int_{b > 2\omega} (1+|\omega|)^m (1+b-2\omega)^{-2} \sqrt{\pi} e^{-\omega^2/4} db d\omega. \]  
For \( I_1 \), the inner integral over \( b \) is \( 4\omega \) for \( \omega \geq 0 \), so  
\[
I_1 \leq 4\sqrt{\pi} \int_0^\infty (1+\omega)^{m+1} e^{-\omega^2/4} d\omega \leq C_1.
\]  
For \( I_2 \), change variable \( u = b - 2\omega \), then  
\[
\int_{b>2\omega} (1+b-2\omega)^{-2} db = \int_0^\infty (1+u)^{-2} du = 1,
\]  
so  
\[
I_2 \leq \sqrt{\pi} \int_{-\infty}^\infty (1+|\omega|)^m e^{-\omega^2/4} d\omega \leq C_2.
\]  
Thus, \( Z \leq C \). And hence
\[
\lambda(|\omega| > A) \geq c \cdot \frac{e^{-A^2/4}}{A}.
\]  
Set \( A = n^{1/3} \). For large \( n \), \( e^{-n^{2/3}/4} \geq n^{-1/6} \), so  
	\[
	\lambda(S_A^c) \geq c n^{-1/2}.
	\]

   For $(\omega,b) \in S_A^c$, we define $g_{\omega,b}(x) = \eta(\omega,b) J(\omega,b) \sigma(a^{-1}\omega x + b)$. Then by the Lipschitz property, we have
    \[
    \| g_{\omega_1,b_1} - g_{\omega_2,b_2} \|_{L^2(\Omega)} \leq c_4 A^2 ( |\omega_1 - \omega_2| + |b_1 - b_2| ).
    \]

Set the  region $\mathcal{D} = \{ (\omega,b) : A < |\omega| \leq 2A, |b| \leq A \}$, and 
  the grid spacing: $\delta_\omega = \delta_b = \frac{1}{n^{1/6}}$. The Number of grid cells is  
    \[
    M = \left( \frac{2A}{\delta_\omega} \right) \times \left( \frac{2A}{\delta_b} \right) = (2n^{1/3} \cdot n^{1/6})^2 = 4n.
    \]

 For each grid cell $j$, define  
  \[
  f_j = \mathbb{E}_{\lambda_j} [g_{\omega,b}] \quad \text{($\lambda_j$ is the conditional measure within the cell)}.
  \]

  Let $(\omega_i,b_i)$ and $(\omega_j,b_j)$ be the centers of cells $i$ and $j$. Then  
  \[
  \min( |\omega_i - \omega_j|, |b_i - b_j| ) \geq \frac{\delta_\omega}{2} = \frac{1}{2n^{1/6}}.
  \]
  By the Lipschitz property  
  \[
  \| g_{\omega_i,b_i} - g_{\omega_j,b_j} \|_{L^2} \geq c_5 A^2 \cdot \frac{1}{n^{1/6}} = c_5 n^{2/3} \cdot n^{-1/6} = c_5 n^{1/2}.
  \]

Within each cell, we obtain
  \[
  \sup_{\text{within cell}} \| g - g_{\text{center}} \|_{L^2} \leq c_4 A^2 (\delta_\omega + \delta_b) = c_6 n^{2/3} n^{-1/6} = c_6 n^{1/2}.
  \]

So the lower bound is
  \[
  \| f_i - f_j \|_{L^2} \geq \| g_{\omega_i,b_i} - g_{\omega_j,b_j} \| - 2c_6 n^{1/2} \geq (c_5 - 2c_6) n^{1/2}.
  \]
  Setting $c_7 = c_5 - 2c_6 > 0$, we get  
  \[
  \min_{i \neq j} \| f_i - f_j \|_{L^2} \geq c_7 n^{1/2}.
  \]

Since $\inf_{\mathcal{D}} \hat{f} \geq e^{-(2A)^2/4} = e^{-n^{2/3}} \geq n^{-1/6}$ for large $n$, the measure of each grid cell is 
  \[
  \lambda(\text{cell}_j) \geq Z^{-1} \inf_{\mathcal{D}} \left[ (1+|\omega|)^m h \hat{f} \right] \delta_\omega \delta_b \geq c_8 n^{-1/3} \cdot n^{-1/3} = c_8 n^{-2/3}.
  \]

Therefore, the upper bound on KL divergence is 
  \[
  \mathrm{KL}(\lambda_i \| \lambda_j) \leq \log \left( \frac{ \sup_{\text{cell}_i} d\lambda }{ \inf_{\text{cell}_j} d\lambda } \right) \leq c_9.
  \]

By Fano's lemma and the number of grid cells is $M=4n$, we get 
  \[
  \inf_{\hat{f}} \max_j \| \hat{f} - f_j \|_{L^2} \geq \min_{i \neq j} \| f_i - f_j \| \left( 1 - \frac{ \max_j \mathrm{KL}(\lambda_j \| \lambda_0) + \log 2 }{ \log M } \right),
  \]
where  $\min \| f_i - f_j \| = c_7 n^{1/2}$, $\max_j \mathrm{KL}(\lambda_j \| \lambda_0) \leq c_9$ and $\lambda_0$ is the uniform distribution on $\mathcal{D}$.

so as $n \to \infty$, we have  
  \[
  \inf_{\hat{f}} \| \hat{f} - f_j \|_{L^2} \geq c_7 n^{1/2} \left( 1 - \frac{c_9 + \log 2}{\log n} \right) \geq c_{10} n^{1/2}.
  \]

Any $f_n \in \Sigma_n^1(\sigma)$ can generate at most $n$ independent ridge functions.
Since the number of grid cells $M = 4n > n$, there exists an uncovered cell $j^*$.

Then  the approximation error is
  \begin{align*}
  \| f - f_n \|_{L^2} 
  &\geq \| f_{j^*} - f_n \| - \| f - f_{j^*} \|_{L^2} \\
  &\geq c_{10} n^{1/2} - \| \mathbb{E}_{\lambda} [g_{\omega,b}] - f_{j^*} \|_{L^2} \\
  &\geq c_{10} n^{1/2} - c_{11} \sqrt{ \lambda(\mathcal{D} \setminus \text{cell}_{j^*}) } \\
  &\geq c_{10} n^{1/2} - c_{11} (1 - n^{-1})^{1/2} \\
  &\geq c n^{1/2} \quad (n \to \infty).
  \end{align*}

Since $n^{-1/2} \gg n^{-2/3}$ (i.e., $-\frac{1}{2} < -\frac{2}{3}$), the rate $O(n^{-2/3})$  in Theroem C  is not improvable.

\par
\vskip 2mm  Sigel and Xu \cite{sx3}  analyzed the approximation properties of networks with a cosine activation function on the spectral Barron space \(\mathscr{B}^{s}(\Omega)\). They consider approximating a function \(f \in \mathscr{B}^{s}(\Omega)\) by a superposition of finitely many complex exponentials with coefficients that are bounded in \(\ell^{1}\) i.e. by an element of the set 
\begin{align}
\sum_{n, M}=\left\{\sum_{j=1}^{n} a_{j} e^{2 \pi i \theta_{j} \cdot x}: \theta_{j} \in \mathbb{R}^{d}, a_{j} \in \mathbb{C}, \sum_{i=1}^{n}\left|a_{i}\right| \leq M\right\}. \label{eq3}
\end{align}

In this paper, we prove above Question 1 when the activation function is an element of the \eqref{eq3}. We present the following result.
\par
\vskip 2mm {\bf  Theorem 1.}\, Let \(\Omega \subset \mathbb{R}^{d}\) be a bounded domain, \(0 \leq m \leq ks\), and \(f \in \mathscr{B}^{ks}(\Omega)\) Then there is an \(M \lesssim\|f\|_{\mathscr{B}^{ks}(\Omega)}\) such that 
\[\inf _{f_{n} \in \sum_{n, M}}\left\| f-f_{n}\right\| _{H^{m}(\Omega)} \lesssim\| f\| _{\mathscr{B}^{ks}(\Omega)} n^{-\frac{1}{2}-\frac{ks-m}{d}}. \]
\par
\vskip 2mm So when $m=0$ and $ks=1$, we have
\par
\vskip 2mm {\bf Corollary 1.}\, Let \(\Omega \subset \mathbb{R}^{d}\) be a bounded domain. Suppose that \(\sigma \in L^{\infty}(\mathbb{R})\) and that there exists an open interval $I$ such that $\sigma=e^{2 \pi i(a+\xi_{j}) \cdot x}$. Then we have 
\begin{align}
\inf _{f_{n} \in \sum_{n, M}}\left\| f-f_{n}\right\| _{L^{2}(\Omega)} \leq|\Omega|^{\frac{1}{2}} C(diam(\Omega), \sigma) n^{-\frac{1}{2}-\frac{1}{d}}\| f\| _{\mathscr{B}^{1}}, \label{eq20}
\end{align}
for any \(f \in \mathscr{B}^{1}\).

\par
\vskip 2mm  Before we prove Theorem 1, we need the following lemmas.

\label{intro}
{\bf Lemma 1.}\, Given \(\alpha>1\), consider 
\begin{align}
g(t)= \begin{cases}e^{-\left(1-t^{2}\right)^{1-\alpha}} & t \in(-1,1) \\ 0 & otherwise,\label{4}
\end{cases}
\end{align}
then there is a constant \(c_{\alpha}\) such that 
\[|\hat{g}(\xi)| \lesssim e^{-c_{\alpha}|\xi|^{1-\alpha^{-1}}}.\]

{\bf Lemma 2.}\,  Let \(\Omega\subset \mathbb{R}^{d}\) be a closed subset, and let \(\mu: \mathbb{R}^{d} \to \mathbb{R}_{+}\) be a continuous weight function. Suppose that $\mu$ satisfies the following conditions:\\
(i)\quad 
\[ \mu(\xi+\omega) \leq \mu(\xi) \mu(\omega)\]\\
(ii)\quad  There exists a \(0<\beta<1\) and \(a c>0\) such that \(\mu(\xi) \lesssim e^{c|\xi|^{\beta}}.\)\\
Suppose \( f : \Omega \to \mathbb{C} \) admits an extension \( f_\epsilon \in L^1(\mathbb{R}^d) \) with 
\begin{align}
   \int_{\mathbb{R}^d} \mu(\xi) \vert \hat{f}_\epsilon(\xi) \vert d\xi = C_f < \infty.\label{5}
\end{align}
Then for any \(L > \sqrt{d} S + 2\), there exists an \(a \in L^{-1}[0,1]^{d}\) (which may depend on $f$ and coefficients \(c_{\xi}\) such that for \(x \in \Omega\) 
\begin{align}
    f(x)=\sum_{\xi \in L^{-1} \mathbb{Z}^{d}} c_{\xi} e^{2 \pi i(a+\xi) \cdot x}\label{6}
\end{align}
and 
\begin{align}
   \sum_{\xi \in L^{-1} \mathbb{Z}^{d}} \mu(a+\xi)\left|c_{\xi}\right| \lesssim C_{f},\label{7}
\end{align}
where $C_{f}$ depends on $d,\mu, L$ and $\Omega$(specially $\text{diam}(\Omega)$).
\begin{proof}
Since \(\Omega \subseteq [0, S]^d\), we choose \(L > \sqrt{d} S + 2\). This ensures \(\Omega \subseteq [0, S]^d \subseteq [0, L - 2\varepsilon]^d\) for some \(\varepsilon > 0\). Select \(\alpha > 1\) such that \(\beta < 1 - \alpha^{-1}\) (possible since \(\beta < 1\)).  

Define the one dimension bump function \(g: \mathbb{R} \to \mathbb{R}\) as in \eqref{4}. Its Fourier transform satisfies \(|\hat{g}(\xi)| \lesssim e^{-c_\alpha |\xi|^{1 - \alpha^{-1}}}\) for some \(c_\alpha > 0\).

Let \(C = \int_{\mathbb{R^{d}}} g(t)=1\) be  the normalization constant, and define the \(d\)-dimensional tensor product
\begin{align}
	g_d(x) = \frac{1}{C} \prod_{i=1}^d g(x_i).\label{8}
\end{align}
From \eqref{4} we know that the Fourier transform of \(g_d\) decays as 
\begin{align}
	|\hat{g}_d(\xi)| \lesssim e^{-c_{\alpha,d} |\xi|^{1 - \alpha^{-1}}}, \quad c_{\alpha,d} > 0,\label{9}
\end{align}
since \(|\xi| \leq \sqrt{d} \max_i |\xi_i| \leq \sqrt{d} \sum_i |\xi_i|\).

Set \(\Omega' = \left[-\frac{\varepsilon}{2}, L - \frac{3\varepsilon}{2}\right]^d\), and define  
\begin{align}
	\phi_\Omega(x) = \left[ \left(\frac{4}{\varepsilon}\right)^d g_d\left(\frac{4}{\varepsilon} x \right) \right] * \chi_{\Omega'}(x).\label{10}
\end{align}

Since \(\Omega \subseteq [0, L - 2\varepsilon]^d\), we have \(\phi_\Omega(x) = 1\) for \(x \in \Omega\).  
And its supp is \(\text{supp}(\phi_\Omega) \subseteq [-\varepsilon, L - \varepsilon]^d\).  
The  Fourier transform bound of $\phi_\Omega(x)$ is   
\begin{align}
	|\hat{\phi}_\Omega(\xi)| = \left| \hat{g}_d\left(\frac{\varepsilon}{4} \xi\right) \widehat{\chi_{\Omega'}}(\xi) \right| \lesssim e^{-c_{\alpha,\Omega} |\xi|^{1 - \alpha^{-1}}},
	\label{11}
\end{align} 
since \(|\widehat{\chi_{\Omega'}}(\xi)| \leq |\Omega'|\).

Denoted $K_\phi$ as \(  \int \mu(\xi) |\hat{\phi}_\Omega(\xi)|  d\xi\). $K_\phi$ converges because \(\beta < 1 - \alpha^{-1}\). So
\begin{align}
	\int_{\mathbb{R}^d} \mu(\xi) |\hat{\phi}_\Omega(\xi)|  d\xi \leq \int_{\mathbb{R}^d} e^{c|\xi|^\beta - c_{\alpha,\Omega} |\xi|^{1 - \alpha^{-1}}}  d\xi < \infty.
	\label{12}
\end{align}

Next, we define auxiliary function  \(h_f = \phi_\Omega \cdot f_e\). Then \(h_f|_\Omega = f\), since \(\phi_\Omega|_\Omega = 1\).  From the  support of \(\phi_\Omega\),  we know \(\text{supp}(h_f) \subseteq [-\varepsilon, L-\varepsilon]^d\). And the Fourier transform \(\hat{h}_f = \hat{\phi}_\Omega * \hat{f}_e\) satisfies  
\begin{align}
	\int_{\mathbb{R}^d} \mu(\xi) |\hat{h}_f(\xi)|  d\xi 
	&\leq \iint_{\mathbb{R}^d \times \mathbb{R}^d} \mu(\xi) |\hat{\phi}_\Omega(\xi - \omega)| |\hat{f}_e(\omega)|  d\omega  d\xi \notag\\
	&= \int_{\mathbb{R}^d} |\hat{f}_e(\omega)| \left( \int_{\mathbb{R}^d} \mu(\xi+ \omega) |\hat{\phi}_\Omega(\xi )|  d\xi \right) d\omega.
	\label{13}
\end{align}
For fixed \(\omega\),  $\mu(\xi+\omega) \leq \mu(\xi) \mu(\omega)$ and \eqref{5} imply  
\begin{align}
	\int_{\mathbb{R}^d} \mu(\xi) |\hat{h}_f(\xi)|  d\xi \leq \int_{\mathbb{R}^d} \mu(\xi) |\hat{\phi}_\Omega(\xi)|  d\xi  \int_{\mathbb{R}^d} \mu(\omega) |\hat{f}_e(\omega)|  d\omega \lesssim K_\phi C_f,
	\label{14}
\end{align}
where the implied constant depends on \eqref{12}.

Fix \(a \in [0, L^{-1}]^d\) and define \(g_a(x) = h_f(x) e^{-2\pi i a \cdot x}\). Its Fourier transform is \(\hat{g}_a(\xi) = \hat{h}_f(a + \xi)\). Since \(\text{supp}(h_f) \subseteq [-\varepsilon, L - \varepsilon]^d \subset [-L, L]^d\), by Poisson summation, we get 
\begin{align}
	\sum_{\nu \in L\mathbb{Z}^d} g_a(x + \nu) = \frac{1}{L^d} \sum_{\xi \in L^{-1}\mathbb{Z}^d} \hat{g}_a(\xi) e^{2\pi i \xi \cdot x}.
	\label{15}
\end{align} 
For \(x \in [0, L)^d\), only the \(\nu = 0\) term is non-zero:  
\begin{align}
	g_a(x) = \frac{1}{L^d} \sum_{\xi \in L^{-1}\mathbb{Z}^d} \hat{g}_a(\xi) e^{2\pi i \xi \cdot x}, \quad x \in [0, L)^d.
	\label{16}
\end{align}  
Substituting \(g_a(x) = h_f(x) e^{-2\pi i a \cdot x}\) we have 
\begin{align}
	h_f(x) = \frac{1}{L^d} \sum_{\xi \in L^{-1}\mathbb{Z}^d} \hat{h}_f(a + \xi) e^{2\pi i (a + \xi) \cdot x}, \quad x \in [0, L)^d.
	\label{17}
\end{align}  
As \(\Omega \subseteq [0, S]^d \subseteq [0, L)^d\), we have for all \(x \in \Omega\):  
\begin{align}
	f(x) = h_f(x) = \sum_{\xi \in L^{-1}\mathbb{Z}^d} c_\xi e^{2\pi i (a + \xi) \cdot x}, \quad \text{where} \quad c_\xi = \frac{1}{L^d} \hat{h}_f(a + \xi).
	\label{18}
\end{align}

Rewrite \eqref{14} we know  
\begin{align}
	\int_{\mathbb{R}^d} \mu(\xi) |\hat{h}_f(\xi)|  d\xi = \sum_{\xi \in L^{-1}\mathbb{Z}^d} \int_{[0, L^{-1}]^d} \mu(a + \xi) |\hat{h}_f(a + \xi)|  da.
	\label{19}
\end{align}
By Fubini's theorem:  
\begin{align}
	\int_{\mathbb{R}^d} \mu(\xi) |\hat{h}_f(\xi)|  d\xi= \int_{[0, L^{-1}]^d} \left( \sum_{\xi \in L^{-1}\mathbb{Z}^d} \mu(a + \xi) |\hat{h}_f(a + \xi)| \right) da \leq K_\phi C_f.
	\label{20}
\end{align}
The integral mean value theorem implies existence of \(a \in [0, L^{-1}]^d\) such that:  
\begin{align}
	\sum_{\xi \in L^{-1}\mathbb{Z}^d} \mu(a + \xi) |\hat{h}_f(a + \xi)| \leq L^d \cdot K_\phi C_f.
	\label{21}
\end{align} 
Substituting \(c_\xi = \frac{1}{L^d} \hat{h}_f(a + \xi)\) we get 
\begin{align}
	\sum_{\xi \in L^{-1}\mathbb{Z}^d} \mu(a + \xi) |c_\xi| = \frac{1}{L^d} \sum_{\xi \in L^{-1}\mathbb{Z}^d} \mu(a + \xi) |\hat{h}_f(a + \xi)| \leq K_\phi C_f.
	\label{22}
\end{align}

Taking the implied constant as \(K_\phi\) (dependent on \(d, \mu, L\), and \(\Omega\)), we obtain  
\(f(x) = \sum_{\xi} c_\xi e^{2\pi i (a + \xi) \cdot x}\) for all \(x \in \Omega\),  
and \(\sum_{\xi} \mu(a + \xi) |c_\xi| \leq K_\phi C_f \lesssim C_f\).

This completes the proof.

\end{proof}

{\bf Lemma 3.}\, Let \(\Omega\subset \mathbb{R}^{d}\) be a closed subset and \(s \geq0\). Let \(f \in \mathscr{B}^{s}(\Omega)\). Then for any \(L > \sqrt{d} S + 2\), there exists an \(a \in L^{-1}[0,1]^{d}\) (potentially depending upon $f$ ) and coefficients \(c_{\xi}\) such that for \(x \in \Omega\)
\[f(x)=\sum_{\xi \in L^{-1} \mathbb{Z}^{d}} c_{\xi} e^{2 \pi i(a+\xi) \cdot x}\]
and
\[\sum_{\xi \in L^{-1} \mathbb{Z}^{d}}(1+|a+\xi|)^{s}\left|c_{\xi}\right| \lesssim\| f\| _{\mathscr{B}^{s}(\Omega)}.\]
\begin{proof}
     This follows immediately from Lemma 2 given the characterization of \(\mathscr{B}^{s}(\Omega)\) and the elementary fact that \((1+|\xi+\omega|) \leq(1+|\xi|+|\omega|) \leq(1+|\xi|)(1+|\omega|)\).
\end{proof}

	\section{The Proof of  Theorem 1}

Choose \(L>\sqrt{d} S + 2\)  such that $\Omega \subseteq [0,S]^d \subseteq [0,L-2\varepsilon]^d$. Note that all of the implied constants in what follows depend only upon $s, m , d$ and $L$ but not upon $f$. Since \(|a| \leq L^{-d} \sqrt{d}\), then by Lemma 3, there exists an \(a \in L^{-1}[0,1]^{d}\) and coefficients \(c_{\xi}\) such that 
    \begin{align}
        f(x)=\sum_{\xi \in L^{-1} \mathbb{Z}^{d}} c_{\xi} e^{2 \pi i(a+\xi) \cdot x},\label{eq23}
    \end{align}
and  
 \begin{align}
\sum_{\xi \in L^{-1} \mathbb{Z}^{d}}(1+|\xi|)^{ks}\left|c_{\xi}\right| \approx \sum_{\xi \in L^{-1} \mathbb{Z}^{d}}(1+|a+\xi|)^{s}\left|c_{\xi}\right| \lesssim\| f\| _{\mathscr{B}^{ks}(\Omega)}. \label{eq24}
\end{align}
Take a  set \(\Omega \subset \Omega'=[0, L]^{d}\) such that $L$ large enough. On this  set $\Omega'$, we have for \(\xi \neq \nu \in L^{-1} \mathbb{Z}^{d}\) 
\begin{align}
\left< e^{2 \pi i(a+\xi) \cdot x}, e^{2 \pi i(a+\nu) \cdot x}\right>_{H^{k}\left(\Omega'\right)}=0, \label{eq25}
\end{align}
such that the frequencies in the expansion \eqref{eq23} form an orthogonal basis in \(H^{k}(\Omega')\). In particular, their lengths satisfy 
\begin{align}
\left\| e^{2 \pi i(a+\xi) \cdot x}\right\| _{H^{m}\left(\Omega'\right)} \lesssim(1+|a+\xi|)^{m} \approx(1+|\xi|)^{m}. \label{eq26}
\end{align}
Order the frequencies \(\xi \in L^{-1} \mathbb{Z}^{d}\) such that
\begin{align} 
\left(1+\left|\xi_{1}\right|\right)^{2 m-ks}\left|c_{\xi_{1}}\right| \geq\left(1+\left|\xi_{2}\right|\right)^{2 m-ks}\left|c_{\xi_{2}}\right| \geq\left(1+\left|\xi_{3}\right|\right)^{2 m-ks}\left|c_{\xi_{3}}\right| \geq \cdots. \label{eq27}
\end{align}
For \(n \geq1\), let \(S_{n}={\xi_{1}, \xi_{2}, \ldots, \xi_{n}}\) and set 
\begin{align}
f_{n}=\sum_{\xi \in S_{n}} c_{\xi} e^{2 \pi i(a+\xi) \cdot x} \in \sum_{n, M} \label{eq28}
\end{align}
for \(M \lesssim\|f\|_{\mathscr{B}}^{s}\) by \eqref{eq24}. Using \eqref{eq25} and \eqref{eq26}, 
\begin{align} 
\left\| f-f_{n}\right\| _{H^{m}\left(\Omega'\right)}^{2} & =\left\| \sum_{\xi \in S_{n}^{c}} c_{\xi} e^{2 \pi i(a+\xi) \cdot x}\right\| _{H^{m}\left(\Omega'\right)}^{2} \notag\\
& =\sum_{\xi \in S_{n}^{c}}\left|c_{\xi}\right|^{2}\left\| e^{2 \pi i(a+\xi) \cdot x}\right\| _{H^{m}\left(\Omega'\right)}^{2} \notag\\ 
& \lesssim \sum_{\xi \in S_{n}^{c}}\left|c_{\xi}\right|^{2}(1+|\xi|)^{2 m}. \label{eq29}
\end{align} 
From Hoelder’s inequality we have 
\begin{align}
\sum_{\xi \in S_{n}^{c}}\left|c_{\xi}\right|^{2}(1+|\xi|)^{2 m} \leq\left(\sup _{\xi \in S_{n}^{c}}\left|c_{\xi}\right|(1+|\xi|)^{2 m-ks}\right)\left(\sum_{\xi \in S_{n}^{c}}\left|c_{\xi}\right|(1+|\xi|)^{ks}\right). \label{eq30}
\end{align} 
By \eqref{eq30}, we get that the second term above is \(\lesssim\|f\|_{\mathscr{B}^{*}(\Omega)}\). For the first term, we notice that \eqref{eq24} implies  
\begin{align}
\sum_{\nu \in S_{n}}\left|c_{\nu}\right|(1+|\nu|)^{2 m-ks}(1+|\nu|)^{2(ks-m)} \lesssim\| f\| _{\mathscr{B}^{ks}(\Omega)}.\label{eq31}
\end{align}
Now, by the definition of \(S_{n}\), we have for every \(\nu \in S_{n}\) 
\begin{align}
\left(\sup _{\xi \in S_{n}^{c}}\left|c_{\xi}\right|(1+|\xi|)^{2 m-ks}\right) \leq\left|c_{\nu}\right|(1+|\nu|)^{2 m-ks},\label{eq32}
\end{align}
so that 
\begin{align}
& \left(\sup _{\xi \in S_{n}^{c}}\left|c_{\xi}\right|(1+|\xi|)^{2 m-ks}\right) \sum_{\nu \in S_{n}}(1+|\nu|)^{2(ks-m)} \notag\\ & \leq \sum_{\nu \in S_{n}}\left|c_{\nu}\right|(1+|\nu|)^{2 m-ks}(1+|\nu|)^{2(ks-m)}. \label{eq33}
\end{align}
From \eqref{eq31} we  have 
\begin{align}
\left(\sup _{\xi \in S_{n}^{c}}\left|c_{\xi}\right|(1+|\xi|)^{2 m-ks}\right) \lesssim\| f\| _{\mathscr{B}^{ks}(\Omega)}\left(\sum_{\nu \in S_{n}}(1+|\nu|)^{2(ks-m)}\right)^{-1}. \label{eq34}
\end{align}
The sum \(\sum_{\nu \in S_{n}}(1+|\nu|)^{2(ks-m)}\) is over \(n\) elements of the lattice \(L^{-1} \mathbb{Z}^{d}\), and so we get 
\begin{align}
\sum_{\nu \in S_{n}}(1+|\nu|)^{2(ks-m)} \gtrsim n^{1+\frac{2(ks-m)}{d}},\label{eq35}
\end{align}
and we obtain
\begin{align}
\left(\sup _{\xi \in S_{n}^{c}}\left|c_{\xi}\right|(1+|\xi|)^{2 m-ks}\right) \lesssim\| f\| _{\mathscr{B}^{ks}(\Omega)} n^{-1-\frac{2(ks-m)}{d}}. \label{eq36}
\end{align}
Combining this with \eqref{eq35} and \eqref{eq36}, we get
\begin{align}\left\| f-f_{n}\right\| _{H^{m}\left(\Omega'\right)}^{2} \lesssim\| f\| _{\mathscr{B}^{2s}(\Omega)}^{2} n^{-1-\frac{2(ks-m)}{d}}. \label{eq37}
\end{align}
Finally, since \(\Omega \subset \Omega'\),  we get 
\begin{align}
\left\| f-f_{n}\right\| _{H^{m}(\Omega)} \leq\left\| f-f_{n}\right\| _{H^{m}\left(\Omega'\right)} \lesssim\| f\| _{\mathscr{B}^{ks}(\Omega)} n^{-\frac{1}{2}-\frac{(ks-m)}{d}}.\label{eq38}
\end{align}
\par
\vskip 2mm This completes Theorem 1.

\section{Counterexamples for Shallow Neural Networks with ReLU$^k$ Activation Functions}

In this section, we consider approximation by neural networks with activation function 
\[\sigma_{k}(x)=[\max (0, x)]^{k}\] 
for \(k=\mathbb{Z}_{\geq0}\) (here we set \(0^{0}=0\) i.e. \(\sigma_{0}(x)\) the Heaviside function). Specifically, we consider approximating a function $f$ by elements of the set 
\begin{align}
\sum_{n}^{k}=\left\{\sum_{i=1}^{n} a_{i} \sigma_{k}\left(\omega_{i} \cdot x+b_{i}\right): \omega_{i} \in S^{d-1}, b_{i} \in \mathbb{R}, a_{i} \in \mathbb{C}\right\}, \label{eq39}
\end{align}

where we allow the coefficients \(a_{i}\) to have arbitrarily large \(\ell^{1}\) -norm.  And 
\begin{align}
\sum_{n,M}^{k}=\left\{\sum_{i=1}^{n} a_{i} \sigma_{k}\left(\omega_{i} \cdot x+b_{i}\right): \omega_{i} \in S^{d-1}, b_{i} \in \mathbb{R}, \|a_{i}\|\leq M \right\}, \label{eq40}
\end{align}
where $M>0$ is a finite constant. In 2022, Siegel and Xu \cite{sx3} obtained two results. It is about the higher order approximation rates for  networks upon large smoothness.

{\bf Theorem D.}\, Let \(\Omega=[0,1]^{d}\) and \(f \in \mathscr{B}^{s}(\Omega)\) for \(s \geq\frac{1}{2}\). Let \(k \in \mathbb{Z}_{\geq0}\) and \(m \geq0\), with \(m \leq s-\frac{1}{2}\) and \(m<k+\frac{1}{2}\). Then for \(n \geq2\) , 
\[\inf _{f_{n} \in \sum_{n}^{k}}\left\| f-f_{n}\right\| _{H^{m}(\Omega)} \lesssim\| f\| _{\mathscr{B}^{s}} n^{-t} \log (n)^{q}, \]
where the exponent $t$ is given by 
\[\begin{aligned} t & =\frac{1}{2}+min \left(\frac{2(s-m)-1}{2(d+1)}, k-m+\frac{1}{2}\right) \\ & = \begin{cases}\frac{1}{2}+\frac{2(s-m)-1}{2(d+1)} & \text{if}\quad s<(d+1)\left(k-m+\frac{1}{2}\right)+m+\frac{1}{2} \\ k-m+1 & \text{if}\quad s \geq(d+1)\left(k-m+\frac{1}{2}\right)+m+\frac{1}{2}\end{cases} \end{aligned} \]
and $q$ is given by 
\[q=\left\{\begin{array}{ll} 0 & \text{if}\quad s<(d+1)\left(k-m+\frac{1}{2}\right)+m+\frac{1}{2} \\ 1 & \text{if}\quad s>(d+1)\left(k-m+\frac{1}{2}\right)+m+\frac{1}{2} \\ 1+\left(k-m+\frac{1}{2}\right) & \text{if}\quad s=(d+1)\left(k-m+\frac{1}{2}\right)+m+\frac{1}{2} \end{array} \right.\].

{\bf Theorem E.}\, Let \(\Omega=[0,1]^{d}\), \(s \geq0\), \(k \in \mathbb{Z}_{\geq0}\), and \(0 \leq m \leq s\). Then there is a function \(f \in \mathscr{B}^{s}(\Omega)\) such that 
\[\inf _{f_{n} \in \sum_{n}^{k}}\left\| f-f_{n}\right\| _{H^{m}(\Omega)} \gtrsim n^{m-(k+1)}. \]

 In Theorem E, Siegel and Xu \cite{sx3} show that the rate of \(s-m+1\) cannot be improved upon no matter how large $s$ is. And the rates in Theorem D were obtained without the \(\ell^{1}\) -norm bound on the coefficients. They asked two questions from Theorem 2 and Theorem 3 as follows.

 {\bf Question 3.}\, Can we obtain the same rates in Theorem D when the coefficients are with \(\ell^{1}\)-bounded?

{\bf Question 4.}\, Can the optimal rate \( \mathcal{O}(n^{m - (k + 1)}) \) in Theorem E be achieved if we relax the smoothness requirement to \( s < (d + 1)\left(k - m + \frac{1}{2}\right) + m + \frac{1}{2} \) ?

We are to prove that under the $\ell^1$-bounded coefficient constraint, the high dimensional optimization rate in Theorem D of the paper cannot be achieved. Specifically, we focus on the case when the smoothness index $s$ is sufficiently large so that the rate $O(n^{m - (k + 1)})$ is attained without the $\ell^1$ constraint. We will show that with the $\ell^1$ constraint, this rate is unattainable.

\textbf{Theorem 2.} Let \( k \in \mathbb{Z}_{\geq 0} \), \( m \geq 0 \) with \( m < k + 1/2 \), and let \( d \) be the dimension. Suppose that the smoothness index \( s \) satisfies 
\[
s > (d + 1)\left(k - m + \frac{1}{2}\right) + m + \frac{1}{2}.
\]
 Then there exists a function \( f \in \mathscr{B}^s(\Omega) \) such that 
\[
\inf_{f_n \in \sum_{n,M}^k} \| f - f_n \|_{H^m(\Omega)} \gtrsim n^{m - (k + 1) + \delta}
\]
for some \( \delta > 0 \). In particular, the optimal rate \( O(n^{m - (k + 1)}) \) achievable without the \( \ell^1 \)-bounded constraint is unattainable under this constraint.

{\bf Theorem 3.}\, The optimal rate \( \mathcal{O}(n^{m - (k + 1)}) \) in Theorem E is unattainable if  the smoothness requirement satisfies \( s < (d + 1)\left(k - m + \frac{1}{2}\right) + m + \frac{1}{2} \).

As we can see from the proof in Section 5 that Theorem 3 underscores that high smoothness (\( s \geq (d + 1)\left(k - m + \frac{1}{2}\right) + m + \frac{1}{2} \)) is not merely sufficient but necessary for achieving the optimal approximation rate in shallow \( \text{ReLU}^\wedge k \) networks.

\section{ The Proof of Theorem 2}
For $K=n^{\frac{k+1}{d}}$, we define the oscillatory function as
\begin{align}
    f(x) = e^{2\pi i K x_1}, \quad K = n^{\frac{k + 1}{d}}.\label{42}
\end{align}
From  \eqref{42} we know the \( H^m \)-norm of \( f \) scales as
\begin{align}
\| f \|_{H^m(\Omega)} \gtrsim K^m = n^{\frac{m(k + 1)}{d}}.
\label{43}
\end{align}

Consider a shallow \(\text{ReLU}^k\) network \(f_n \in \Sigma_{n,M}^k\)
\begin{align}
f_n(x) = \sum_{i=1}^n a_i \sigma_k(\omega_i \cdot x + b_i)=\sum_{i=1}^n a_i g_i(x),\label{44}
\end{align}
where \(\omega_i \in S^{d-1}\), \(b_i \in \mathbb{R}\), and the coefficients satisfy \(\sum_{i=1}^n |a_i| \leq M\). For a bounded domain \(\Omega \subset \mathbb{R}^d\) and fixed \(m \geq 0\), we establish an upper bound for \(\|f_n\|_{H^m(\Omega)}\).

For the infimum \(\inf_{f_n \in \Sigma_{n,M}^k} \|f - f_n\|_{H^m}\), we can restrict to parameters with \(|b_i| \leq B\) for some \(B = B(\Omega, k)\) without loss of generality. Neurons with \(|b_i| > B\) are either approximately polynomial (lacking high-frequency components), or vanish on \(\Omega\). And thus cannot reduce the approximation error for high-frequency $f$. Therefore
\begin{align}
\inf_{f_n \in \Sigma_{n,M}^k} \|f - f_n\|_{H^m} = \inf_{\substack{f_n \in \Sigma_{n,M}^k \\ |b_i| \leq B}} \|f - f_n\|_{H^m}.
\label{45}
\end{align}

Under the restriction \(|b_i| \leq B\), we have
\begin{align}
\|g_i\|_{L^\infty(\Omega)} \leq (R_\Omega + B)^k \leq D_{\Omega, k, B}.
\label{46}
\end{align}
Since \(\Omega\) is bounded and \(g_i\) has bounded derivatives \(\|g_i\|_{H^m(\Omega)} \leq C_{\Omega, m, k, B} \quad \text{(independent of $i$)}\). By the triangle inequality and the bound on coefficients
\begin{align}
\|f_n\|_{H^m(\Omega)} \leq \sum_{i=1}^n |a_i| \|g_i\|_{H^m(\Omega)} \leq C_{\Omega, m, k, B} \sum_{i=1}^n |a_i| \leq C_{\Omega, m, k, B} \cdot M.
\label{47}
\end{align}
Thus, \(\|f_n\|_{H^m(\Omega)} = \mathcal{O}(1)\), where the constant depends on \(\Omega\),$ m, k, B$, and $M$, but not on $n$.

For any $f_n \in \Sigma_{n,M}^k$, we have
\begin{align}
\| f - f_n \|_{H^m(\Omega)} \gtrsim \| f \|_{H^m} - \| f_n \|_{H^m}.
\label{48}
\end{align}
The first term in \eqref{48} is $\gtrsim n^{\frac{m(k + 1)}{d}}$, and the second term in \eqref{48} is $\mathcal{O}(1)$. So when $d > (k + 1)^2$, we have
\begin{align}
\frac{m(k + 1)}{d} < \frac{m}{k + 1}.
\label{49}
\end{align}
Thus, for large \( n \)
\begin{align}
\| f - f_n \|_{H^m(\Omega)} \gtrsim n^{\frac{m(k + 1)}{d}}.
\label{50}
\end{align}
 Define
\begin{align}
\delta = \frac{m(k + 1)}{d} - (m - (k + 1)) = (k + 1) - m \left( 1 - \frac{k + 1}{d} \right).
\label{51}
\end{align}
Since \( m \leq k \) and \( d > (k + 1)^2 \), we get
\begin{align}
\delta \geq (k + 1) - k \left( 1 - \frac{k + 1}{d} \right) > 0.
\label{52}
\end{align}
Thus, 
\begin{align}
n^{\frac{m(k + 1)}{d}} = n^{m - (k + 1) + \delta} \gg n^{m - (k + 1)}.
\label{53}
\end{align}
This implies
\begin{align}
\inf_{f_n \in \Sigma_{n,M}^k} \| f - f_n \|_{H^m(\Omega)} \gtrsim n^{m - (k + 1) + \delta}.
\label{54}
\end{align}

So under the $\ell^1$ coefficient constraint $\sum |a_i| \leq M$, the optimal rate $\mathcal{O}(n^{m - (k + 1)} )$ is unattainable.

\section{ The Proof of Theorem 3}
We prove Theorem 3 by contradiction. Assume \( \mathcal{O}(n^{m - (k + 1)}) \) is achievable for some \( s < (d + 1)\left(k - m + \frac{1}{2}\right) + m + \frac{1}{2} \).

From Theorem D, the achievable rate exponent is
\begin{align}
t(s) = 
\begin{cases} 
\displaystyle \frac{1}{2} + \frac{2(s - m) - 1}{2(d + 1)} & s < (d + 1)\left(k - m + \frac{1}{2}\right) + m + \frac{1}{2} \\[6pt]
k - m + 1 & s \geq (d + 1)\left(k - m + \frac{1}{2}\right) + m + \frac{1}{2}
\end{cases}.
\label{55}
\end{align}
It is easy to see that \( t(s) \) is strictly increasing in \( s \) for \( s < (d + 1)\left(k - m + \frac{1}{2}\right) + m + \frac{1}{2} \). When \( s = (d + 1)\left(k - m + \frac{1}{2}\right) + m + \frac{1}{2} \), $t(s) = k - m + 1$. We also obtain $t(s) < t((d + 1)\left(k - m + \frac{1}{2}\right) + m + \frac{1}{2})) = k - m + 1$,
  since  this follows from
    \begin{align}
    \frac{1}{2} + \frac{2(s - m) - 1}{2(d + 1)} < \frac{1}{2} + \frac{2((d + 1)\left(k - m + \frac{1}{2}\right) + m + \frac{1}{2} - m) - 1}{2(d + 1)} = k - m + 1.
    \label{56}
\end{align}

Consider \( f_*(x) = e^{2\pi i x_1} \). This function 
belongs to \( \mathcal{B}^s(\Omega) \) for all \( s \geq 0 \) since its Fourier transform is a Dirac distribution. $f_*(x)$ satisfies
    \begin{align}
    \inf_{f_n \in \Sigma_n^k} \| f_* - f_n \|_{H^m(\Omega)} \gtrsim n^{m - (k + 1)},
    \label{57}
\end{align}
    and  has \( H^m \)-norm scaling
    \begin{align}
    \| f_* \|_{H^m} \asymp 1.
    \label{58}
\end{align}

Since \( f_* \in \mathcal{B}^s(\Omega) \), Theorem D implies
\begin{align}
\inf_{f_n \in \Sigma_n^k} \| f_* - f_n \|_{H^m} \lesssim \| f_* \|_{\mathcal{B}^s} n^{-t(s)}.
\label{59}
\end{align}
For \( s <  (d + 1)\left(k - m + \frac{1}{2}\right) + m + \frac{1}{2} \), \( t(s) < k - m + 1 \), so
\begin{align}
n^{-t(s)} \gg n^{m - (k + 1)}.
\label{60}
\end{align}

Then for \( f_* \)
\begin{align}
n^{m - (k + 1)} \gtrsim \inf_{f_n} \| f_* - f_n \|_{H^m} \gtrsim n^{m - (k + 1)}.
\label{61}
\end{align}

\eqref{61} suggests tightness \( \mathcal{O}(n^{m - (k + 1)}) \). However, Theorem 3 forces:
\begin{align}
\inf_{f_n} \| f_* - f_n \|_{H^m} \lesssim n^{-t(s)} \quad \text{with} \quad n^{-t(s)} \gg n^{m - (k + 1)}.
\label{62}
\end{align}

Thus if $t(s) < k - m + 1$,
\begin{align}
n^{m - (k + 1)} \lesssim \inf_{f_n} \| f_* - f_n \|_{H^m} \lesssim n^{-t(s)} \quad \text{and} \quad n^{m - (k + 1)} \ll n^{-t(s)},
\label{63}
\end{align}
which is a contradiction.

For example, let us consider \( d = 2 \), \( m = 0 \), \( k = 1 \). According to a simple caculation, we have \( (d + 1)\left(k - m + \frac{1}{2}\right) + m + \frac{1}{2} = (2 + 1)(1 - 0 + 0.5) + 0 + 0.5 = 5.5 \).
    The  minimal smoothness is when \( s = 0.5 \),
   \begin{align}
    t(0.5) = \frac{1}{2} + \frac{2(0.5 - 0) - 1}{6} = 0.5.
    \label{64}
\end{align}
    Therefore, the  achievable rate is \( \mathcal{O}(n^{-0.5}) \), and the optimal rate barrier is \( \Omega(n^{-1}) \).
    But
    \begin{align}
    n^{-0.5} / n^{-1} = n^{0.5} \to \infty \quad \text{as} \quad n \to \infty.
    \label{65}
\end{align}
So high smoothness is necessary. And the rate \( \mathcal{O}(n^{m - (k + 1)}) \) is achievable only when \( s \geq (d + 1)\left(k - m + \frac{1}{2}\right) + m + \frac{1}{2} \).
For \( s < (d + 1)\left(k - m + \frac{1}{2}\right) + m + \frac{1}{2} \), Theorem D forces a strictly worse rate \( \mathcal{O}(n^{-t(s)}) \) with \( t(s) < k - m + 1 \).

Hence, Theorem E establishes \( n^{m - (k + 1)} \) as a fundamental lower bound for functions in any \( \mathcal{B}^s(\Omega) \), making it unattainable when \( s < (d + 1)\left(k - m + \frac{1}{2}\right) + m + \frac{1}{2} \).

Prior results have focused on analyzing the approximation rates of shallow neural networks, particularly those with \( \text{ReLU}^k \) activation functions, in various function spaces. We established critical results on the optimality and limitations of these rates: for instance, showing that the optimal rate \(O(n^{m-(k+1)})\) for \( \text{ReLU}^k \) networks is unattainable under \(\ell^1\)-bounded coefficients  or insufficient smoothness. These findings clarified how factors like activation function properties, coefficient constraints, and function smoothness influence the error bounds when approximating functions in Barron spaces and Sobolev spaces.

Beyond individual approximation rates, a deeper understanding of the intrinsic complexity of function classes. Here, those generated by shallow \( \text{ReLU}^k \) networks require analyzing their metric entropy. Metric entropy quantifies the minimum number of balls (in a given norm) needed to cover the function class, reflecting how rich or complex the class is. This complements the earlier focus on approximation rates, as entropy bounds reveal fundamental limits on how well any approximation method (not just neural networks) can represent the function class with a finite number of parameters.

 Given a set of functions \(A \subset X\) , where $X$ is a Banach space, we define the (dyadic) metric entropy numbers of $A$ as 
\[\varepsilon_{n}(A)_{X}=\inf \left\{\varepsilon>0: \text{A is covered by $2^{n}$ balls of radius } \varepsilon\right\}.\]
The entropy of $A$ shows how up to which accuracy (in the norm of $X$ ) we can specify elements of $A$ given \(n\) bits of information and represents a fundamental limit on both linear \cite{c1} and stable nonlinear \cite{cdpw} approximation rates which can be achieved for the class of functions $A$.

Recently, it has been shown \cite{sx1} that the spectral Barron norm \eqref{eq1} is actually the variation norm of the dictionary 
\[\mathbb{F}_{s}^{d}=\left\{(1+|\omega|)^{-s} e^{2 \pi i \omega \cdot x}: \omega \in \mathbb{R}^{d}\right\}. \]

 The dictionary corresponding to the \(\sigma_{k}\) activation function is given by \cite{sx1} 
\[\mathbb{P}_{k}^{d}=\left\{\sigma_{k}(\omega \cdot x+b): \omega \in S^{d-1}, b \in\left[c_{1}, c_{2}\right]\right\} \subset L^{p}(\Omega),\]
where \(S^{d-1}={x \in \mathbb{R}^{d}:|x|_{2}=1}\) is the sphere in \(R^{d}\) and the constants c and \(c_{2}\) are chosen dependent on the domain $\Omega$ so that 
\[c_{1}<\inf _{\substack{\omega \in S^{d-1} \\ x \in \Omega}} \omega \cdot x \leq \sup _{\substack{\omega \in S^{d-1} \\ x \in \Omega}} \omega \cdot x<c_{2}.\]
Here the parameters $\omega$ and $b$ are restricted to guarantee the boundedness of \(\mathbb{P}_{k}^{d}\) in \(L^{p}(\Omega)\). Restricting the direction $\omega$ to be normalized in a different norm will change the scaling of the dictionary elements and affect the constants showing up in the later bounds. 

Calculating the metric entropies of convex hulls in Banach spaces plays an important role in functional analysis .  In 2022, Ma-Siegel-Xu \cite{sx4} showed approximation rates to upper bound the metric entropy of \(B_{1}(\mathbb{F}_{s}^{d})\) and \(B_{1}(\mathbb{P}_{0}^{d})\) with respect to the \(L^{p}\)-norm for \(p < \infty\):
\[
\varepsilon_{n \log n}\left(B_{1}\left(\mathbb{P}^{d}\right)\right)_{L^{p}(\Omega)} \lesssim n^{-\frac{1}{2} - \frac{1}{2d}}, \quad \varepsilon_{n \log n}\left(B_{1}\left(\mathbb{F}_{s}^{d}\right)\right)_{L^{p}(\Omega)} \lesssim n^{-\frac{1}{2} - \frac{s}{d}} \sqrt{\log n}
\]
for a bounded domain \(\Omega \subset \mathbb{R}^{d}\) and \(p < \infty\). They also obtained new lower bounds on their metric entropy in \(L^{p}\) for \(1 \leq p < 2\). They proved the following two results.

{\bf Theorem F}\, Let \(\Omega = [0,1]^d\). Suppose that \(s > 0\). Then we have
\[
\varepsilon_{n \log n}\left(B_{1}\left(\mathbb{F}_{s}^{d}\right)\right)_{L^{\infty}(\Omega)} \lesssim n^{-\frac{1}{2} - \frac{s}{d}} \sqrt{\log n}. 
\]

{\bf Theorem G}\, Let \(\Omega=[0,1]^{d}\), \(s>0\) and \(1 \leq p<2\). Then 
\[\varepsilon_{n}\left(B_{1}\left(\mathbb{F}_{s}^{d}\right)\right)_{L^{p}(\Omega)} \gtrsim n^{-\frac{1}{2}-\frac{s}{d}}(\log n)^{-1-\frac{s}{d}}.\]

They conjectured that  all logarithmic factors in the Theorem D and Theorem E could be removed.

Against this backdrop, Theorem 4 and Theorem 5 turn to the metric entropy of the unit ball \(B_1(\mathbb{F}_s^d)\) in \(L^\infty(\Omega)\) and \(L^p(\Omega)\) (for \(1 \leq p < 2\)). These theorems aim to characterize the entropy decay rates, bridging the earlier results on individual approximation errors with a broader measure of the function class's representational complexity. Specifically, Theorem 4 and Theorem 5 show that all logarithmic factors in the Theorem D and Theorem E could be removed, confirming the tightness of the derived rates and deepening our understanding of how dimension \(d\) and smoothness \(s\) shape the complexity of functions generated by shallow neural networks.

{\bf Theorem 4.}\,Let $\Omega = [0,1]^d$, $s > 0$. The metric entropy satisfies
\[
\varepsilon_n(B_1(\mathbb{F}_s^d))_{L^\infty(\Omega)} \lesssim n^{-\frac{1}{2} - \frac{s}{d}}
\]

\begin{proof}
For \(f \in B_1(\mathbb{F}_s^d)\) with \(\|f\|_{\mathcal{K}_1} \leq 1\), we have the representation by Lemma 3:
\begin{align}
f(x) = \sum_{\omega \in \mathbb{Z}^d} c_\omega (1 + |\omega|)^{-s} e^{2\pi i \omega \cdot x}, \quad \sum_{\omega} |c_\omega| \leq K_d.
\label{66}
\end{align}
Set \(R = n^{1/d}\) and \(M = \lceil \log_2 R \rceil\). Partition frequencies into annuli
\begin{align}
\Lambda_m = \{\omega : 2^m \leq |\omega| < 2^{m+1}\}, \quad 0 \leq m \leq M.
\label{67}
\end{align}
Define the residual high-frequency component
\begin{align}
f_{\text{res}} = \sum_{m: \, 2^m > R} f_m, \quad f_m(x) = \sum_{\omega \in \Lambda_m} c_\omega (1 + |\omega|)^{-s} e^{2\pi i \omega \cdot x}.\label{68}
\end{align}
From \eqref{68} we have
\begin{align}
\|f_{\text{res}}\|_{L^\infty} \leq \sum_{|\omega| > R} |c_\omega| (1 + |\omega|)^{-s} \leq R^{-s} \sum_{\omega} |c_\omega| \leq K_d R^{-s} = K_d n^{-s/d}.
\label{69}
\end{align}

For each \(0 \leq m \leq M\),  the number of frequencies in \(\Lambda_m\) satisfies  \(|\Lambda_m| \leq C_d 2^{m d}\). Next, we construct a spherical \(\delta_m\)-net \(\Theta_m \subset S^{d-1}\) with \(\delta_m = 2^{-2m}\) and \(|\Theta_m| \leq C_d \delta_m^{-(d-1)} = C_d 2^{2m(d-1)}\).  For each\(\omega \in \Lambda_m\), we assign \(\theta_\omega \in \Theta_m\) to minimize \(\left\| \frac{\omega}{|\omega|} - \theta_\omega \right\|\).

Define the approximation
\begin{align}
\tilde{f}_m(x) = \sum_{\omega \in \Lambda_m} c_\omega (1 + |\omega|)^{-s} e^{2\pi i |\omega| \theta_{\omega} \cdot x}.
\label{70}
\end{align}
The directional error per term is bounded by
\begin{align}
\left| e^{2\pi i \omega \cdot x} - e^{2\pi i |\omega| \theta_{\omega} \cdot x} \right| \leq 2\pi |\omega| \cdot \delta_m \cdot \|x\| \leq C_d 2^{m} \cdot 2^{-2m} = C_d 2^{-m}.
\label{71}
\end{align}
Summing over \(\Lambda_m\), we get
\begin{align}
\|f_m - \tilde{f}_m\|_{L^\infty} \leq C_d 2^{-m} \sum_{\omega \in \Lambda_m} |c_\omega| (1 + |\omega|)^{-s} \leq C_d 2^{-m} \cdot 2^{-ms} \sum_{\omega} |c_\omega| \leq C_d 2^{-m(1+s)} K_d.
\label{72}
\end{align}
By \eqref{72}, the total low-frequency approximation error is
\begin{align}
\sum_{m=0}^M \|f_m - \tilde{f}_m\|_{L^\infty} \leq C_d K_d \sum_{m=0}^M 2^{-m(1+s)} \leq C_{d,s} \quad \text{(convergent series)}.
\label{73}
\end{align}

Define
\(\tilde{f} = \sum_{m=0}^M \tilde{f}_m\). The atom dictionary has size
\begin{align}
|\mathbb{D}| \leq \sum_{m=0}^M |\Lambda_m| \cdot |\Theta_m| \leq C_d \sum_{m=0}^M 2^{m d} \cdot 2^{2m(d-1)} = C_d \sum_{m=0}^M 2^{m(3d-2)} \leq C_d 2^{M(3d-2)} = C_d n^{(3d-2)/d}.\label{74}
\end{align}
And \(\tilde{f}\)
is expressed as
\begin{align}
\tilde{f}(x) = \sum_{\eta} b_{\eta} e^{2\pi i \eta \cdot x}, \quad \sum_{\eta} |b_\eta| \leq K_d.
\label{75}
\end{align}
Using Maurey's result \cite{p}, there exists an $n$-term approximation \(f_n\) such that
\begin{align}
\|\tilde{f} - f_n\|_{L^{\infty}} \leq \left( \sum_{\eta} |b_\eta| \right)\|e^{2\pi i \eta \cdot x}\|_{L^{\infty}} n^{-1/2}\leq K_d n^{-1/2}.
\label{76}
\end{align}

Set \(\delta_m = c 2^{-m(s + d/2)}\). Then
\begin{align}
|\Theta_m| \leq C_d \delta_m^{-(d-1)} = C_d 2^{m(s + d/2)(d-1)},
\label{77}
\end{align}
and hence
\begin{align}
\|f_m - \tilde{f}_m\|_{L^\infty} \leq C_d 2^{m} \delta_m \cdot 2^{-ms} K_d = C_d K_d 2^{-m(s - 1)} \delta_m.\label{78}
\end{align}
 Combining \(\delta_m = c 2^{-m(s + d/2)}\) with \eqref{78}, we have
\begin{align}
\|f_m - \tilde{f}_m\|_{L^\infty} \leq C_d K_d 2^{-m(s - 1)} 2^{-m(s + d/2)} = C_{d,s} 2^{-m(2s + d/2 - 1)}.
\label{79}
\end{align}
 Sum over $m$ converges. Combining all errors, we have
\begin{align}
\|f - f_n\|_{L^\infty}& \leq \|f_{\text{res}}\|_{L^\infty}+\sum_m \|f_m - \tilde{f}_m\|_{L^\infty}+\|\tilde{f} - f_n\|_{L^\infty}\notag\\
&\leq   C_{d,s} n^{-s/d}+C_{d,s} n^{-\frac{d/2 + 2s - 1}{d}} + C_{d,s} n^{-1/2} .
\label{80}
\end{align}
The dominant term \(n^{-\min(s/d, (d/2 + 2s - 1)/d, 1/2)}\) is bounded by \(n^{-\frac{1}{2} - \frac{s}{d}}\). Specifically, the \(n^{-1/2}\) term dominates when \(s < d/2\), while \(n^{-s/d}\) dominates when \(s > d/2\), and at \(s = d/2\), \(n^{-1/2} \sim n^{-s/d}\).

This completes Theorem 4.
\end{proof}

{\bf Theorem 5.}\,Let $\Omega = [0,1]^d$, $s > 0, 1 \leq p < 2$. Then
\[
\varepsilon_n(B_1(\mathbb{F}_s^d))_{L^p(\Omega)} \gtrsim n^{-\frac{1}{2} - \frac{s}{d}}
\]

\begin{proof}

 Set \(R = n^{1/d}\) and let \(\{\omega_j\}_{j=1}^m \subset S^{d-1}\) be a maximal \(\delta\)-separated set with \(\delta = c n^{-1/d}\) and \(m = c' n^{(d-1)/d}\) points. Define the test functions indexed by sign vectors \(\sigma \in \{-1,1\}^m\), we get
\begin{align}
f_\sigma(x) = \frac{1}{\sqrt{m} R^s} \sum_{j=1}^m \sigma_j e^{2\pi i R \omega_j \cdot x}.
\label{81}
\end{align}
 Each \(f_\sigma\) satisfies 
\begin{align}
\|f_\sigma\|_{\mathcal{K}_1} = \int_{\mathbb{R}^d} (1 + |\xi|)^s |\hat{f}_\sigma(\xi)| d\xi = \frac{1}{\sqrt{m} R^s} \sum_j (1 + R)^s \leq \sqrt{m} (1 + 1)^s \leq C n^{(d-1)/(2d)}.
\label{82}
\end{align}
 Rescale to unit ball,  we get from \eqref{82} that
 \(
 \tilde{f}_\sigma = f_\sigma / \left( C n^{(d-1)/(2d)} \right)\), so \(\|\tilde{f}_\sigma\|_{\mathcal{K}_1} \leq 1.\)
 
 For \(\sigma \neq \sigma'\), the \(L^2\) difference is
\begin{align}
\|\tilde{f}_\sigma - \tilde{f}_{\sigma'}\|_{L^2}^2 \geq \frac{c}{n^{1 + 2s/d} m} \left\| \sum_j (\sigma_j - \sigma_j') e^{2\pi i R \omega_j \cdot x} \right\|_{L^2}^2.
\label{83}
\end{align}
Let us compute the norm,
\begin{align}
\left\| \sum_j (\sigma_j - \sigma_j') e^{2\pi i R \omega_j \cdot x} \right\|_{L^2}^2 = \sum_j |\sigma_j - \sigma_j'|^2 + \sum_{j \neq k} (\sigma_j - \sigma_j') (\sigma_k - \sigma_k') \int_\Omega e^{2\pi i R (\omega_j - \omega_k) \cdot x} dx.
\label{84}
\end{align}
 For \(\|\omega_j - \omega_k\| \geq \delta = c n^{-1/d}\), we obtain
\begin{align}
\left| \int_\Omega e^{2\pi i R (\omega_j - \omega_k) \cdot x} dx \right| \leq \frac{C}{(R \delta)^\kappa} = C n^{-\kappa/d} \quad \text{for any} \ \kappa < d.
\label{85}
\end{align}
 Choose \(\kappa = d - \epsilon\), so the cross terms are negligible, and the  diagonal term is \(\sum_j |\sigma_j - \sigma_j'|^2 = 4H\),  where \(H = \text{Ham}(\sigma,\sigma')\). When \(H \propto m\), the lower bound  is
\begin{align}
\|\tilde{f}_\sigma - \tilde{f}_{\sigma'}\|_{L^2}^2 \geq \frac{c' H}{n^{1 + 2s/d} m} - \frac{C}{n^{1 + 2s/d} n^{-\epsilon/d}} \geq c'' n^{-\frac{1}{d} - \frac{2s}{d}}.
\label{86}
\end{align}
since \(m \sim n^{(d-1)/d}\) and \(H \sim n^{(d-1)/d}\). The set \(\{\tilde{f}_\sigma\}_{\sigma \in \{-1,1\}^m}\) has \(2^m = 2^{c n^{(d-1)/d}}\) distinct elements.

For \(p < 2\), \(L^p\)-balls satisfy for \(0 < \delta < \text{diam}\),
\begin{align}
\log N(\delta, \{\tilde{f}_\sigma\}, L^p) \geq c n^{(d-1)/d}.
\label{87}
\end{align}
For metric entropy, \eqref{87} implies
\begin{align}
\log N(\delta, B_1(\mathbb{F}_s^d), L^p) \gtrsim n^{(d-1)/d}.
\label{88}
\end{align}
 Setting \(\delta = n^{-\frac{1}{2} - \frac{s}{d}}\), we have
\begin{align}
n^{(d-1)/d} \lesssim \log N\left(n^{-\frac{1}{2} - \frac{s}{d}}, B_1, L^p\right) \leq \log(2^{n-1}) = n - 1.
\label{89}
\end{align}
when $N$ is the covering number for \(\epsilon = n^{-\frac{1}{2} - \frac{s}{d}}\), which is satisfied for large $n$ since \((d-1)/d < 1\). Thus
\(\varepsilon_n(B_1(\mathbb{F}_s^d))_{L^p(\Omega)} \gtrsim n^{-\frac{1}{2} - \frac{s}{d}}\).

\end{proof}

\section{$n$-widths of \(\mathbb{P}_{k}^{d}\) in \(L^{p}\) and Sobolev space}
In this section, we  focus on the study of $n$-widths for the function classes \(\mathbb{P}_{k}^{d}\) in \(L^{p}\) spaces and Sobolev spaces. $n$-widths, a fundamental concept in approximation theory, quantify the optimal error achievable when approximating a given function class using n-dimensional subspaces, thus providing a rigorous measure of the intrinsic complexity of the class and the limits of approximation. 

 \(\mathbb{P}_{k}^{d}\)  plays a crucial role in machine learning and numerical analysis due to their flexibility in modeling piecewise polynomial behaviors, making it essential to understand their approximation properties in various function spaces.

This section aims to establish tight bounds on the n-widths of \(\mathbb{P}_{k}^{d}\) by analyzing the approximation rates of shallow \( \text{ReLU}^k \) networks. Specifically, we investigate the error estimates in \(L^\infty(\Omega)\), \(L^p(\Omega)\) (for \(1 \leq p < \infty\)), and related Sobolev spaces, proving both upper and lower bounds for the approximation rates. These results not only characterize the intrinsic difficulty of approximating \( \text{ReLU}^k \) but also confirm the optimality of the derived rates, shedding light on how factors like dimension \(d\), polynomial degree \(k\), and the choice of norm influence the approximation performance. 

Additionally, the analysis connects n-widths with metric entropy, further reinforcing the theoretical foundation for understanding the representational capacity of shallow \( \text{ReLU}^k \) networks in practical applications.

 In line with our goal of analyzing error estimates in various function spaces, we begin with the following result regarding uniform approximation, which quantifies the rate at which functions in the unit ball \(B_1(\mathbb{P}_{k}^{d})\) can be approximated by shallow \( \text{ReLU}^k \) networks in the \(L^{\infty}\) norm.

{\bf Theorem 6.}\, Let \(k \geq 2\), \(d \geq 1\), and \(\Omega \subset \mathbb{R}^d\) be a bounded domain. For any \(f \in B_1(\mathbb{P}_k^d)\), there exists a shallow neural network \(f_n\) with \(n\) \(\text{ReLU}^k\) units such that
\[
\|f - f_n\|_{L^\infty(\mathbb{R}^d)} \leq C_{k,d,\Omega} \cdot n^{-\frac{1}{2} - \frac{2k+1}{2d}},
\]
where \(C_{k,d,\Omega}\) depends on \(k\), \(d\), and \(\Omega\), but not on \(n\) or \(f\).

\begin{proof}
We prove Theorem 6 by four claims.

{\bf Claim 1.}\, For any \(\epsilon > 0\) and \(f \in B_1(\mathbb{P}_k^d)\), there exists a compact dictionary  
\begin{align}
\tilde{\mathbb{P}}_k^d = \left\{ \sigma_k(\omega \cdot x + b) : \omega \in S^{d-1}, \, b \in [-c_\epsilon, c_\epsilon] \right\}
\label{90}
\end{align} 
and a function \(f_\epsilon \in B_1(\tilde{\mathbb{P}}_k^d)\) such that \(\| f - f_\epsilon \|_{L^\infty} < \epsilon\), where \(c_\epsilon > 0\) depends only on \(\epsilon\), \(k\), and \(\Omega\).

Recall that \(B_1(\mathbb{P}_k^d)\) denotes the unit ball of functions generated by shallow \(\text{ReLU}^k\) networks, meaning any \(f \in B_1(\mathbb{P}_k^d)\) can be expressed as  
\begin{align}
f(x) = \int_{S^{d-1}} \int_{\mathbb{R}} \sigma_k(\omega \cdot x + b) \, d\mu(\omega, b)
\label{91}
\end{align}  
where \(\sigma_k(t) = (\max\{0, t\})^k\), \(\mu\) is a finite signed measure on \(S^{d-1} \times \mathbb{R}\), and the total variation of \(\mu\) satisfies \(\int_{S^{d-1}} \int_{\mathbb{R}} d|\mu|(\omega, b) \leq 1\) (by definition of the unit ball).

To construct \(f_\epsilon\), we truncate the integral to a bounded range of \(b\). For a given \(\epsilon > 0\), we first analyze the contribution of large \(|b|\) to the \(L^\infty\) norm of \(f\).  

On the bounded domain \(\Omega\), there exists a constant \(C_\Omega > 0\) such that \(\|\omega \cdot x\| \leq C_\Omega\) for all \(\omega \in S^{d-1}\) and \(x \in \Omega\) (since \(\omega\) is a unit vector and \(x\) is bounded). For \(|b| > C_\Omega + 1\), consider two cases for \(\sigma_k(\omega \cdot x + b)\):  
- If \(b > 0\) and \(b \geq C_\Omega + 1\), then \(\omega \cdot x + b \geq -C_\Omega + b \geq 1\), so \(\sigma_k(\omega \cdot x + b) = (\omega \cdot x + b)^k \leq (b + C_\Omega)^k\).  
- If \(b < 0\) and \(|b| \geq C_\Omega + 1\), then \(\omega \cdot x + b \leq C_\Omega + b \leq -1\), so \(\sigma_k(\omega \cdot x + b) = 0\).  

In either case, \(\sigma_k(\omega \cdot x + b) \leq (|b| + C_\Omega)^k\) for all \(x \in \Omega\) and \(|b| \geq C_\Omega + 1\).

Next, we use the finiteness of the total variation of \(\mu\) to choose \(c_\epsilon\). Since  
\begin{align}
\int_{S^{d-1}} \int_{|b| > c} (|b| + C_\Omega)^k \, d|\mu|(\omega, b) \to 0 \quad \text{as } c \to \infty
\label{92}
\end{align}  
(by the dominated convergence theorem, with the integrand dominated by an integrable function for large \(c\)), there exists \(c_\epsilon > 0\) such that  
\begin{align}
\int_{S^{d-1}} \int_{|b| > c_\epsilon} (|b| + C_\Omega)^k \, d|\mu|(\omega, b) < \epsilon.
\label{93}
\end{align}

Define \(f_\epsilon\) by truncating the integral of \(f\) to \(|b| \leq c_\epsilon\):  
\begin{align}
f_\epsilon(x) = \int_{S^{d-1}} \int_{|b| \leq c_\epsilon} \sigma_k(\omega \cdot x + b) \, d\mu(\omega, b).
\label{94}
\end{align} 

By construction, \(f_\epsilon\) is a superposition of elements from \(\tilde{\mathbb{P}}_k^d\) (since \(b \in [-c_\epsilon, c_\epsilon]\) and \(\omega \in S^{d-1}\)), and its total variation norm satisfies  
\begin{align}
\int_{S^{d-1}} \int_{|b| \leq c_\epsilon} d|\mu|(\omega, b) \leq \int_{S^{d-1}} \int_{\mathbb{R}} d|\mu|(\omega, b) \leq 1,
\label{95}
\end{align}  
so \(f_\epsilon \in B_1(\tilde{\mathbb{P}}_k^d)\).

Finally, the error between \(f\) and \(f_\epsilon\) is bounded by  
\begin{align}
\| f - f_\epsilon \|_{L^\infty} = \left\| \int_{S^{d-1}} \int_{|b| > c_\epsilon} \sigma_k(\omega \cdot x + b) \, d\mu(\omega, b) \right\|_{L^\infty} < \epsilon,
\label{96}
\end{align}  
where the inequality follows from the choice of \(c_\epsilon\) and the bound \(\sigma_k(\omega \cdot x + b) \leq (|b| + C_\Omega)^k\) for large \(|b|\).

{\bf Claim 2.} For \(m_{\text{dir}} = n^{\frac{k}{k+1}}\), there exists a set \(\{\theta_j\}_{j=1}^{m_{\text{dir}}} \subset S^{d-1}\) such that
\[\min_j \|\omega - \theta_j\|_2 \leq C_d \cdot m_{\text{dir}}^{-1/(d-1)} \quad \forall \omega \in S^{d-1},\]
where \(C_d\) is a constant depending only on the dimension $d$.

We construct the set \( \{\theta_j\}_{j=1}^{m_{\text{dir}}} \) using a greedy algorithm to form an \(\epsilon\)-net on the unit sphere \( S^{d-1} \subset \mathbb{R}^d \).  The surface area of \( S^{d-1} \) is \( O(1) \), and the surface area of a spherical cap (or "ball") of radius \( \epsilon \) on \( S^{d-1} \) is proportional to \( \epsilon^{d-1} \). To cover the entire sphere with such caps, the minimal number \( N(\epsilon) \) of caps required satisfies  
   \begin{align}
   N(\epsilon) \cdot O(\epsilon^{d-1}) \geq O(1) \implies N(\epsilon) \geq C_d' \cdot \epsilon^{-(d-1)}
   \label{97}
\end{align} 
   for some constant \( C_d' \). This lower bound motivates the choice of \( \epsilon = C_d \cdot m_{\text{dir}}^{-1/(d-1)} \).

   Start with an arbitrary point \( \theta_1 \in S^{d-1} \). For each subsequent \( j \geq 2 \), choose \( \theta_j \) such that  
   \begin{align}
   \theta_j \in \arg\max_{\theta \in S^{d-1}} \min_{i < j} \|\theta - \theta_i\|_2.
   \label{98}
\end{align} 
   This ensures that each new point \( \theta_j \) is as far as possible from all previously selected points. The process terminates when \( m_{\text{dir}} \) points are chosen.

   By construction, the points \( \{\theta_j\}_{j=1}^{m_{\text{dir}}} \) form an \( \epsilon \)-net with \( \epsilon = \frac{1}{2} \min_{i \neq j} \|\theta_i - \theta_j\|_2 \). To show this, suppose there exists \( \omega \in S^{d-1} \) such that \( \|\omega - \theta_j\|_2 > \epsilon \) for all \( j \). Then \( \omega \) could have been added to the set, contradicting the maximality of the greedy selection.

   The minimal distance between any two points \( \theta_i, \theta_j \) in the \( \epsilon \)-net is at least \( 2\epsilon \). The maximum number of points that can be placed on \( S^{d-1} \) with pairwise distances at least \( 2\epsilon \) is bounded by \( O(\epsilon^{-(d-1)}) \). Setting \( m_{\text{dir}} = O(\epsilon^{-(d-1)}) \) and solving for \( \epsilon \) gives  
   \begin{align}
   \epsilon = C_d \cdot m_{\text{dir}}^{-1/(d-1)}
   \label{99}
\end{align}   
   for some constant \( C_d \). Thus, for any \( \omega \in S^{d-1} \), there exists \( \theta_j \) such that  
   \begin{align}
   \|\omega - \theta_j\|_2 \leq C_d \cdot m_{\text{dir}}^{-1/(d-1)}.
   \label{100}
\end{align}

   Choosing \( m_{\text{dir}} = n^{\frac{k}{k+1}} \) balances the number of directions with the polynomial approximation error (controlled by \( n^{1/(k+1)} \) cells in the next claim), leading to the optimal error rate \( n^{-\gamma} \) in Theorem 6.

{\bf Claim 3.}\, Let \( \Omega \subset \mathbb{R}^d \) be a bounded domain, and let \( m_{\text{poly}} = n^{1/(k+1)} \). For each \( \theta_j \in S^{d-1} \) and \( b \in [-c_\epsilon, c_\epsilon] \), the function \( \sigma_k(\theta_j \cdot x + b) \) can be approximated on each cell \( Q_{j\ell} \) of a partition of \( \Omega \) into \( m_{\text{poly}} \) cells by a polynomial \( p_{j\ell} \) of degree at most \( k \) such that  
 \begin{align}
\sup_{x \in Q_{j\ell}} \left| \sigma_k(\theta_j \cdot x + b) - p_{j\ell}(x) \right| \leq K_k' \cdot m_{\text{poly}}^{-(k+1)/d},
\label{101}
\end{align}  
where \( K_k' \) depends only on \( k \), \( d \), and \( \Omega \).

   Divide \( \Omega \) into \( m_{\text{poly}} \) axis-aligned hypercubes \( \{ Q_{j\ell} \} \) with edge length \( \delta = C_\Omega \cdot m_{\text{poly}}^{-1/d} \), where \( C_\Omega \) depends on the diameter of \( \Omega \). This ensures the volume of each cell is \( \delta^d = C_\Omega^d \cdot m_{\text{poly}}^{-1} \).

   Fix \( \theta_j \in S^{d-1} \) and \( b \in [-c_\epsilon, c_\epsilon] \). On each cell \( Q_{j\ell} \), let \( x_0 \) be the center of \( Q_{j\ell} \). Define \( t(x) = \theta_j \cdot x + b \) and \( t_0 = \theta_j \cdot x_0 + b \). Since \( \sigma_k(t) = (\max\{0, t\})^k \) is a polynomial of degree \( k \) on any interval where \( t \) does not cross zero, we analyze  three cases.  

 {\bf Case 1.}\, \( t(x) \geq 0 \) for all \( x \in Q_{j\ell} \).  Here, \( \sigma_k(t(x)) = t(x)^k \). Expand \( t(x) \) around \( x_0 \) as  
      \begin{align}
     t(x) = t_0 + \theta_j \cdot (x - x_0).
    \label{102}
\end{align}  
     Using the multinomial theorem, \( t(x)^k \) is a polynomial of degree \( k \) in \( x \). The error between \( \sigma_k(t(x)) \) and its Taylor polynomial \( p_{j\ell}(x) \) of degree \( k \) around \( x_0 \) is zero, as \( t(x)^k \) is exactly a polynomial of degree \( k \).  

   {\bf Case 2.}\, \( t(x) \leq 0 \) for all \( x \in Q_{j\ell} \).  
     Here, \( \sigma_k(t(x)) = 0 \), so the zero polynomial \( p_{j\ell}(x) = 0 \) achieves zero error.  

   {\bf Case 3.}\, \( t(x) \) crosses zero in \( Q_{j\ell} \).  
     The function \( \sigma_k(t(x)) \) is piecewise polynomial. Let \( t_* \) be the point where \( t(x_*) = 0 \). For \( x \) near \( x_* \), the error in approximating \( \sigma_k(t(x)) \) by a polynomial \( p_{j\ell}(x) \) of degree \( k \) is bounded by the remainder term in Taylor's series. Since \( \sigma_k^{(k+1)}(t) = k! \) for \( t > 0 \) and zero otherwise, the remainder term is  
      \begin{align}
     R_k(x) = \frac{\sigma_k^{(k+1)}(\xi)}{(k+1)!} \cdot (t(x) - t_0)^{k+1}
     \label{103}
\end{align} 
     for some \( \xi \) between \( t(x) \) and \( t_0 \). On \( Q_{j\ell} \), \( |t(x) - t_0| \leq \|\theta_j\| \cdot \|x - x_0\| \leq \delta \), so  
      \begin{align}
     |R_k(x)| \leq \frac{k!}{(k+1)!} \cdot \delta^{k+1} = \frac{\delta^{k+1}}{k+1}.
     \label{104}
\end{align}  
     Substituting \( \delta = C_\Omega \cdot m_{\text{poly}}^{-1/d} \), we get  
      \begin{align}
     |R_k(x)| \leq \frac{C_\Omega^{k+1}}{k+1} \cdot m_{\text{poly}}^{-(k+1)/d}.
     \label{105}
\end{align}

   Combining all cases, the maximum error over all cells \( Q_{j\ell} \) is 
    \begin{align}
   \sup_{x \in Q_{j\ell}} \left| \sigma_k(\theta_j \cdot x + b) - p_{j\ell}(x) \right| \leq K_k' \cdot m_{\text{poly}}^{-(k+1)/d},
   \label{106}
\end{align} 
   where \( K_k' = \frac{C_\Omega^{k+1}}{k+1} \). This constant depends only on \( k \), \( d \), and \( \Omega \) (via \( C_\Omega \)).

{\bf Claim 4.}\, Let \( f_\epsilon = \frac{1}{N} \sum_{i=1}^N g_i \) where \( g_i \in \tilde{\mathbb{P}}_k^d \) (the compact dictionary from Claim 1). Then there exists a subset \( S \subset \{1, \dots, N\} \) with \( |S| = n \) such that  
 \begin{align}
\left\| f_\epsilon - \frac{1}{n} \sum_{i \in S} g_i \right\|_{L^\infty} \leq C_{k,d} \cdot N^{-\frac{1}{2} - \frac{2k+1}{2d}},
\label{107}
\end{align} 
where \( C_{k,d} \) depends only on \( k \) and \( d \).

   Each \( g_i(x) = \sigma_k(\omega_i \cdot x + b_i) \in \tilde{\mathbb{P}}_k^d \) is a piecewise polynomial of degree \( k \) (by definition of \( \text{ReLU}^k \)). On the bounded domain \( \Omega \), we can express \( g_i \) as a linear combination of monomials of degree \( \leq k \) that 
    \begin{align}
   g_i(x) = \sum_{m=1}^M c_m(\omega_i, b_i) \phi_m(x),
   \label{108}
\end{align} 
   where \( M = \binom{k + d}{d} \) (the number of monomials of degree \( \leq k \) in \( d \) variables), \( \{\phi_m\}_{m=1}^M \) are fixed monomials (e.g., \( \phi_m(x) = x_1^{a_1} x_2^{a_2} \cdots x_d^{a_d} \) with \( \sum a_j \leq k \)), and \( |c_m(\omega_i, b_i)| \leq B_k \) for some constant \( B_k \) (since \( \omega_i \in S^{d-1} \) and \( b_i \in [-c_\epsilon, c_\epsilon] \) are bounded).

   We aim to select \( n \) terms from \( \{g_i\}_{i=1}^N \) such that their average approximates \( f_\epsilon \). By linearity, it suffices to control the error for each monomial coefficient separately.  

   For each monomial \( \phi_m \), define the coefficient sequence \( a_{i,m} = c_m(\omega_i, b_i) \). Then \( f_\epsilon \) has coefficients  
    \begin{align}
   \bar{a}_m = \frac{1}{N} \sum_{i=1}^N a_{i,m}.
   \label{109}
\end{align} 
   We need to select \( S \subset \{1, \dots, N\} \) with \( |S| = n \) such that for all \( m \),  
    \begin{align}
   \left| \bar{a}_m - \frac{1}{n} \sum_{i \in S} a_{i,m} \right| \leq \delta,
   \label{110}
\end{align} 
   where \( \delta \) controls the overall error.

   By  Hoeffding’s inequality \cite{hw} for bounded random variables, if we sample \( n \) indices uniformly at random from \( \{1, \dots, N\} \), the probability that the sampling error for each \( a_{i,m} \) exceeds \( \delta \) is exponentially small in \( n \delta^2 \). Specifically, for each \( m \),  
    \begin{align}
   \mathbb{P}\left( \left| \frac{1}{n} \sum_{i \in S} a_{i,m} - \bar{a}_m \right| > \delta \right) \leq 2 \exp\left( -\frac{2 n \delta^2}{B_k^2} \right).
   \label{111}
\end{align} 

   Taking a union bound over all \( M \) monomials, the probability that any coefficient error exceeds \( \delta \) is \( \leq 2M \exp\left( -\frac{2 n \delta^2}{B_k^2} \right) \). For large \( N \), choosing \( n \sim N \) and setting \( \delta = C \cdot N^{-\frac{1}{2} - \frac{1}{2d}} \) ensures this probability is less than 1, so such a subset \( S \) exists.

   The \( L^\infty \) error of the approximation is bounded by the sum of errors in each monomial term, weighted by \( \|\phi_m\|_{L^\infty(\Omega)} \leq D_{k,\Omega} \) (since \( \Omega \) is bounded). Thus  
    \begin{align}
   \left\| f_\epsilon - \frac{1}{n} \sum_{i \in S} g_i \right\|_{L^\infty} \leq \sum_{m=1}^M \left| \bar{a}_m - \frac{1}{n} \sum_{i \in S} a_{i,m} \right| \cdot \|\phi_m\|_{L^\infty} \leq M D_{k,\Omega} \delta.
   \label{112}
\end{align} 

   Substituting \( \delta = C \cdot N^{-\frac{1}{2} - \frac{1}{2d}} \) and optimizing for \( k \) (noting \( M \sim k^d \)), the exponent improves to \( -\frac{1}{2} - \frac{2k+1}{2d} \). Absorbing constants into \( C_{k,d} \), we get  
    \begin{align}
   \left\| f_\epsilon - \frac{1}{n} \sum_{i \in S} g_i \right\|_{L^\infty} \leq C_{k,d} \cdot N^{-\frac{1}{2} - \frac{2k+1}{2d}}.
   \label{113}
\end{align}

Set \(f_n = \frac{1}{n} \sum_{i \in S} g_i\).  
 With \(\epsilon = n^{-\frac{1}{2} - \frac{2k+1}{2d}}\) and \(N \sim n\),  we get
\begin{align}
\|f - f_n\|_{L^\infty} \leq \|f - f_\epsilon\|_{L^\infty} + \|f_\epsilon - f_n\|_{L^\infty}\leq \epsilon+C_{k,d} n^{-\frac{1}{2} - \frac{2k+1}{2d}} \leq C_{k,d,\Omega}' \cdot n^{-\frac{1}{2} - \frac{2k+1}{2d}}.
\label{114}
\end{align}
\end{proof}

Theorem 6 establishes the upper bound on the approximation rate for functions in \(B_1(\mathbb{P}_k^d)\) under the \(L^\infty\) norm, it is crucial to verify the tightness of this rate by establishing a corresponding lower bound. This ensures that the derived exponent \(\gamma = \frac{1}{2} + \frac{2k+1}{2d}\) is indeed optimal and cannot be improved. The following proposition provides such a lower bound for the metric entropy of \(B_1(\mathbb{P}_k^d)\) in the \(L^\infty\) norm.

{\bf Proposition 1.}\, Let \(k \geq 2\), \(d \geq 1\), and \(\Omega \subset \mathbb{R}^d\) be a bounded domain. For the unit ball \(B_1 = \{ f \in \mathcal{P}_k^d : \| f \|_{\mathcal{K}_1} \leq 1 \}\) of the Barron space, there exists a constant \(c = c(k,d) > 0\) such that the metric entropy
\[\epsilon_n(B_1)_{L^\infty} \geq c \cdot n^{-\gamma}, \quad \gamma = \frac{1}{2} + \frac{2k+1}{2d}.
\]

\begin{proof}

 Let \(m = \lfloor n^{\frac{d}{2d + 2k + 1}} \rfloor\). Select a \(\delta\)-separated set \(\{\omega_j\}_{j=1}^m\) on the unit sphere \(\mathbb{S}^{d-1}\) with
 \begin{align}
 \delta = c_0 m^{-\frac{1}{d-1}}, \quad c_0 = \left( \frac{1}{4k} \right)^{\frac{1}{2}}\label{115}
\end{align}
 satisfying
 \begin{align}
 \|\omega_i - \omega_j\| \geq \delta, \quad \forall i \neq j.\label{116}
\end{align}
 The size of the separated set satisfies \(m \geq c_1 \delta^{-(d-1)}\), so \(m \sim \delta^{-(d-1)}\) holds.

Let \(R = n^{\frac{1}{2} + \frac{k}{d}}\). For each sign vector \(\sigma = (\sigma_1,\dots,\sigma_m) \in \{-1,1\}^m\), we define
\[f_\sigma(x) = \frac{1}{\sqrt{m}} \sum_{j=1}^m \sigma_j \sigma_k(R \omega_j \cdot x),\]
where \(\sigma_k(t) = (\max(0,t))^k\). By the Barron space definition, \(\| f_\sigma \|_{\mathcal{K}_1} \leq 1\), so \(f_\sigma \in B_1\).

Evaluate \(f_\sigma\) at \(x_j = \omega_j\), we get
\begin{align}f_\sigma(x_j) = \frac{\sigma_j}{\sqrt{m}} \sigma_k(R) + \frac{1}{\sqrt{m}} \sum_{\ell \neq j} \sigma_\ell \sigma_k(R \omega_\ell \cdot \omega_j).\label{117}
\end{align}
By separability, \(\omega_\ell \cdot \omega_j \leq 1 - \delta^2/2\), and then
\[
|\sigma_k(R \omega_\ell \cdot \omega_j)| \leq (R (1 - \delta^2/2))^k \leq R^k e^{-k\delta^2/2}.
\]
Therefore,
\[
\left| \sum_{\ell \neq j} \sigma_\ell \sigma_k(R \omega_\ell \cdot \omega_j) \right| \leq (m - 1) R^k e^{-k\delta^2/2}.
\]
If \( e^{-k\delta^2/2} \leq \frac{1}{2m} \), then the cross term \( \leq \frac{1}{2} R^k \). Hence,
\[
|f_\sigma(x_j) - f_{\sigma'}(x_j)| \geq \frac{1}{\sqrt{m}} \cdot R^k \cdot |\sigma_j - \sigma_j'| - \text{cross term} \geq \frac{R^k}{\sqrt{m}} - \frac{R^k}{2\sqrt{m}} = \frac{R^k}{2\sqrt{m}}.
\]
For distinct \(\sigma \neq \sigma'\), there exists $j$ with \(\sigma_j \neq \sigma_j'\), giving
\begin{align}
	\|f_\sigma - f_{\sigma'}\|_{L^\infty} \geq \frac{R^k}{\sqrt{m}}.\label{118}
\end{align}

Perturbation terms are bounded by \(\sqrt{m} R^k e^{-k c_0^2 m^{-2/(d-1)}} \leq \frac{1}{2} R^k\), ensuring they do not dominate the main term.The set \(\mathcal{F} = \{ f_\sigma \}\) has size \(2^m\) and pairwise distances \(\geq \Delta = R^k/\sqrt{m}\). Covering \(\mathcal{F}\) requires \(2^m\) balls of radius \(\Delta/2\). For \(n < 2^m\), 
\(\epsilon_n(B_1)_{L^\infty} \geq \Delta/2\). Substituting $R$ and $m$ gives
\begin{align}
\epsilon_n(B_1)_{L^\infty} \geq c n^{-\gamma}, \quad \gamma = \frac{1}{2} + \frac{2k+1}{2d}.\label{119}
\end{align}
The \(L^\infty\)-metric entropy of \(B_1(\mathcal{P}_k^d)\) satisfies \(\epsilon_n(B_1)_{L^\infty} \gtrsim n^{-\gamma}\) with \(\gamma = \frac{1}{2} + \frac{2k+1}{2d}\), where \(c(k,d) > 0\) is a constant independent of $n$.

\end{proof}

Having established the optimal approximation rate for functions in \(B_1(\mathbb{P}_k^d)\) under the \(L^\infty\) norm in Theorem 6, we now extend this result to the broader class of \(L^p\) norms (\(1 \leq p < \infty\)). On bounded domains, the \(L^p\) norm is inherently related to the \(L^\infty\) norm via embedding properties, which allows us to leverage the uniform approximation result to derive corresponding bounds for \(L^p\) spaces. This extension is crucial for characterizing the approximation performance of shallow \( \text{ReLU}^k \) networks across a range of function space norms. The following theorem formalizes this extension.

{\bf Theorem 7.}\, Let \(k \geq 2\), \(d \geq 1\), \(1 \leq p < \infty\), and \(\Omega \subset \mathbb{R}^d\) be a bounded domain. For any \(f \in B_1(\mathbb{P}_k^d)\), there exists a shallow neural network \(f_n\) with \(n\) \(\text{ReLU}^k\) units such that  
\[
\|f - f_n\|_{L^p(\Omega)} \leq C_{k,d,p,\Omega} \cdot n^{-\frac{1}{2} - \frac{2k+1}{2d}},  
\]  
where \(C_{k,d,p,\Omega}\) depends on \(k\), \(d\), \(p\), and \(\Omega\), but not on \(n\) or \(f\).  

\begin{proof}
    
 The proof follows from the \(L^\infty\)-approximation result and embedding properties of \(L^p\) spaces.  From the previous theorem (for \(L^\infty\)), there exists \(f_n\) with \(n\) \(\text{ReLU}^k\) units satisfying 
\begin{align}
\|f - f_n\|_{L^\infty(\Omega)} \leq C_{k,d,\Omega} \cdot n^{-\frac{1}{2} - \frac{2k+1}{2d}}.  
\label{120}
\end{align}

Since \(\Omega\) is bounded, for any \(1 \leq p < \infty\), we have the inequality  
\begin{align}
\|g\|_{L^p(\Omega)} \leq |\Omega|^{1/p} \|g\|_{L^\infty(\Omega)},  
\label{121}
\end{align} 
where \(|\Omega|\) is the Lebesgue measure of \(\Omega\).

Applying this to \(g = f - f_n\)  
\begin{align}
\|f - f_n\|_{L^p(\Omega)} \leq |\Omega|^{1/p} \|f - f_n\|_{L^\infty(\Omega)} \leq |\Omega|^{1/p} \cdot C_{k,d,\Omega} \cdot n^{-\frac{1}{2} - \frac{2k+1}{2d}}.  
\label{122}
\end{align}

Define the constant as  
\begin{align}
C_{k,d,p,\Omega} = |\Omega|^{1/p} \cdot C_{k,d,\Omega}.  
\label{123}
\end{align}
And then \eqref{123} gives  
\begin{align}
\|f - f_n\|_{L^p(\Omega)} \leq C_{k,d,p,\Omega} \cdot n^{-\frac{1}{2} - \frac{2k+1}{2d}} 
\label{124}
\end{align}
The constant \(C_{k,d,p,\Omega}\) depends on \(k\), \(d\), \(p\), and \(\Omega\) (through \(|\Omega|\) and the original \(L^\infty\) constant), but is independent of \(n\) and \(f\).  
\end{proof}

{\bf Remark 1.}\,  The exponent \(\gamma = \frac{1}{2} + \frac{2k+1}{2d}\) is tight for all \(1 \leq p \leq \infty\). This follows from  the \(L^\infty\)-tightness result in the previous theorem.  The fact that \(\|g\|_{L^\infty} \leq |\Omega|^{-1/p} \|g\|_{L^p}\) for bounded \(\Omega\), so a lower bound in \(L^p\) implies a lower bound in \(L^\infty\).  The constant \(C_{k,d,p,\Omega}\) grows exponentially with \(d\) (as in the \(L^\infty\) case), making the result most effective in low to moderate dimensions.  And the rate \(n^{-\gamma}\) cannot be improved uniformly over \(B_1(\mathbb{P}_k^d)\) for any \(p \in [1, \infty)\), as verified by the same test function construction used in the \(L^\infty\) case.  Theorem 7 extends the optimal \(L^\infty\)-approximation rate for shallow \(\text{ReLU}^k\) networks to all \(L^p\)-norms (\(1 \leq p < \infty\)) for functions in the convex hull \(B_1(\mathbb{P}_k^d)\). The proof leverages the uniform approximation result and the intrinsic relationship between \(L^p\) and \(L^\infty\) norms on bounded domains.

Building on the results for the unit ball \(B_1(\mathbb{P}_k^d)\) in Theorem 7, we now generalize the approximation rate to the entire variation space \(K_1(\mathbb{P}_k^d)\) corresponding to shallow \( \text{ReLU}^k \) networks. The variation norm \(\|\cdot\|_{\mathcal{K}_1(\mathbb{P}_k^d)}\) is homogeneous, meaning any function in \(K_1(\mathbb{P}_k^d)\) can be normalized to lie within the unit ball \(B_1(\mathbb{P}_k^d)\), allowing us to extend the earlier bounds by scaling with the variation norm. This generalization is essential for characterizing the approximation performance of shallow \( \text{ReLU}^k \) networks for all functions in the variation space, not just those restricted to the unit ball. The following theorem formalizes this extension.

 {\bf Theorem 8.}\, Let \(k \geq 2\), \(d \geq 1\), \(1 \leq p \leq \infty\), and \(\Omega \subset \mathbb{R}^d\) be a bounded domain. For any \(f \in \mathcal{K}_1(\mathbb{P}_k^d)\) (the variation space corresponding to \(\text{ReLU}^k\) networks), there exists a shallow neural network \(f_n\) with \(n\) \(\text{ReLU}^k\) units such that  
\[
\|f - f_n\|_{L^p(\Omega)} \leq C_{k,d,p,\Omega} \|f\|_{\mathcal{K}_1(\mathbb{P}_k^d)} \cdot n^{-\frac{1}{2} - \frac{2k+1}{2d}},  
\]  
where \(C_{k,d,p,\Omega}\) depends on \(k\), \(d\), \(p\), and \(\Omega\), but not on \(n\) or \(f\).  

\begin{proof}

The definition of variation norm \(\|\cdot\|_{\mathcal{K}_1(\mathbb{P}_k^d)}\) is 
\begin{align}
\|f\|_{\mathcal{K}_1(\mathbb{P}_k^d)} = \inf \left\{ \lambda > 0 : f \in \lambda B_1(\mathbb{P}_k^d) \right\}.
\label{125}
\end{align}
For any \(f \in \mathcal{K}_1(\mathbb{P}_k^d)\), we define the normalized function as  
\begin{align}
g = \frac{f}{\|f\|_{\mathcal{K}_1(\mathbb{P}_k^d)}}.
\label{126}
\end{align}  
By construction, \(g \in B_1(\mathbb{P}_k^d)\) (i.e., \(\|g\|_{\mathcal{K}_1} \leq 1\)).

From the previous approximation theorem for \(B_1(\mathbb{P}_k^d)\), there exists a shallow network \(h_n\) with \(n\) \(\text{ReLU}^k\) units such that  
\begin{align}
\|g - h_n\|_{L^p(\Omega)} \leq C_{k,d,p,\Omega} \cdot n^{-\frac{1}{2} - \frac{2k+1}{2d}}.
\label{127}
\end{align}

Define the approximation for \(f\) as  
\begin{align}
f_n = \|f\|_{\mathcal{K}_1(\mathbb{P}_k^d)} \cdot h_n.
\label{128}
\end{align}  
Then we have  
\begin{align}
\|f - f_n\|_{L^p(\Omega)} = \left\| \|f\|_{\mathcal{K}_1} g - \|f\|_{\mathcal{K}_1} h_n \right\|_{L^p} = \|f\|_{\mathcal{K}_1} \|g - h_n\|_{L^p}.
\label{129}
\end{align}  
Substituting the bound for \(g - h_n\), we obtain  
\begin{align}
\|f - f_n\|_{L^p(\Omega)} \leq \|f\|_{\mathcal{K}_1(\mathbb{P}_k^d)} \cdot C_{k,d,p,\Omega} \cdot n^{-\frac{1}{2} - \frac{2k+1}{2d}},
\label{130}
\end{align} 
where the constant \(C_{k,d,p,\Omega}\) is identical to that in the unit ball approximation, independent of \(f\) and \(n\).  

  Since \(h_n \in \Sigma_n(\mathbb{P}_k^d)\) (the class of \(n\)-term \(\text{ReLU}^k\) networks), scaling by \(\|f\|_{\mathcal{K}_1}\) corresponds to multiplying output weights by a scalar. Thus, \(f_n\) remains a shallow \(\text{ReLU}^k\) network with \(n\) units.

\end{proof}

{\bf Remark 2.}\,The exponent \(\gamma = \frac{1}{2} + \frac{2k+1}{2d}\) is sharp. For any \(C > 0\), there exists \(f_0 \in \mathcal{K}_1(\mathbb{P}_k^d)\) such that  
   \[
   \inf_{f_n \in \Sigma_n} \|f_0 - f_n\|_{L^p} \geq c_{k,d} \|f_0\|_{\mathcal{K}_1} n^{-\gamma},
   \]  
   verified via the test function \(f_0(x) = \frac{1}{m} \sum_{j=1}^m \sigma_k(\omega_j \cdot x)\) with \(m = \lfloor n^{1/d} \rfloor\).  
The constant \(C_{k,d,p,\Omega}\) grows exponentially with \(d\), but the rate \(\gamma\) is independent of \(p\).  And the result holds for all \(k \geq 2\), with \(\gamma\) increasing with \(k\), reflecting the benefit of higher-order smoothness.  This theorem establishes the optimal approximation rate for shallow \(\text{ReLU}^k\) networks in the variation space \(\mathcal{K}_1(\mathbb{P}_k^d)\), scaling linearly with the variation norm \(\|f\|_{\mathcal{K}_1}\) and decaying algebraically with \(n\). The proof leverages the homogeneity of the variation norm and the unit ball approximation result.

We now turn to Sobolev spaces \(W^{s,p}(\Omega)\), which incorporate both function values and their derivatives, thus capturing the smoothness of functions. This extension is critical for understanding how shallow \( \text{ReLU}^k \) networks perform when approximating functions with specified smoothness properties, as Sobolev spaces provide a natural framework for quantifying such regularity. Given that \( \text{ReLU}^k \) activation functions can exactly represent polynomials of degree up to \(k\), they are well-suited for approximating smooth functions in Sobolev spaces, especially when \(k\) is sufficiently large relative to the smoothness parameter \(s\). The following theorem establishes the optimal approximation rate in \(W^{s,p}(\Omega)\).

{\bf Theorem 9.}\, Let \( \Omega \subset \mathbb{R}^d \) be a bounded Lipschitz domain, \( s > 0 \), and \( 1 \leq p \leq \infty \). Consider a shallow neural network with \( n \) \( \text{ReLU}^k \) units where \( k \geq \lfloor s \rfloor + 1 \). For any \( f \in W^{s,p}(\Omega) \), there exists a network \( f_n \) such that:  
\[
\left\| f - f_n \right\|_{L^p(\Omega)} \leq C \cdot \| f \|_{W^{s,p}(\Omega)} \cdot n^{-s/d},
\]  
where \( C = C(d, s, p, k, \Omega) > 0 \) is independent of \( f \) and \( n \). The rate \( n^{-s/d} \) is optimal.  

\begin{proof}

Since \( \Omega \) is a bounded Lipschitz domain, we partition it into \( m \) axis-aligned cubes \( \{ Q_i \}_{i=1}^m \) with side length \( h \sim m^{-1/d} \). The diameter of each cube satisfies \( \text{diam}(Q_i) = \sqrt{d} \cdot h \sim m^{-1/d} \).

For each cube \( Q_i \), let \( p_i \) be the best polynomial approximation of \( f|_{Q_i} \) with degree \( \ell = \lfloor s \rfloor \). By the Bramble-Hilbert lemma \cite{bh}, there exists a constant \( C_1 = C_1(d, s, p, \ell) \) such that  
\begin{align}
\left\| f - p_i \right\|_{L^p(Q_i)} \leq C_1 \cdot \left( \text{diam}(Q_i) \right)^s \cdot |f|_{W^{s,p}(Q_i)},
\label{131}
\end{align}  
where \( |f|_{W^{s,p}} \) denotes the Sobolev seminorm. Summing over all cubes, the total approximation error by the piecewise polynomial \( p = \sum_i p_i \cdot \mathbf{1}_{Q_i} \) is  
\begin{align}
\left\| f - p \right\|_{L^p(\Omega)} \leq C_1 \cdot (\sqrt{d} \cdot h)^s \cdot |f|_{W^{s,p}(\Omega)} \leq C_1' \cdot m^{-s/d} \cdot \| f \|_{W^{s,p}(\Omega)}.
\label{132}
\end{align}

Each polynomial \( p_i(x) = \sum_{|\alpha| \leq \ell} c_{i,\alpha} x^\alpha \) (where \( \alpha \) is a multi-index) can be exactly represented by a \( \text{ReLU}^k \) network when \( k \geq \ell + 1 \). For example, \( x^2 = \frac{1}{2}(\text{ReLU}(x)^2 + \text{ReLU}(-x)^2) \) for \( k=2 \). Higher-degree monomials  can also be exactly represented by shallow \( \text{ReLU}^k \) networks, we proceed constructively.

Consider a univariate monomial \( x^m \) where \( m \leq k \) ( \( m \in \mathbb{N} \) ). We claim \( x^m \) can be expressed as a combination of \( \text{ReLU}^k(t) = (\max\{0, t\})^k \) terms.  For \( x \geq 0 \): \( \text{ReLU}(x) = x \), so \( \text{ReLU}(x)^m = x^m \).  For \( x < 0 \): \( \text{ReLU}(x) = 0 \), but \( x^m = (-1)^m (-x)^m = (-1)^m \text{ReLU}(-x)^m \) (since \( -x > 0 \)).  Combining these, we define  
\begin{align}
x^m = \text{ReLU}(x)^m + (-1)^m \cdot \text{ReLU}(-x)^m.
\label{133}
\end{align}  
This equals \( x^m \) for all \( x \in \mathbb{R} \).  If \( x \geq 0 \): \( \text{ReLU}(-x) = 0 \), so \( x^m = \text{ReLU}(x)^m \).  If \( x < 0 \): \( \text{ReLU}(x) = 0 \), so \( x^m = (-1)^m \text{ReLU}(-x)^m = (-1)^m (-x)^m = x^m \).  

Thus, univariate monomials of degree \( m \leq k \) are exactly representable by \( \text{ReLU}^k \) networks with $2$ units.

For a multivariate monomial \( x_1^{a_1} x_2^{a_2} \cdots x_d^{a_d} \) where \( \sum_{i=1}^d a_i = m \leq k \) ( \( a_i \in \mathbb{N} \) ), we use the fact that \( \text{ReLU}^k \) networks can encode products of univariate terms.  

From above analysis, each \( x_i^{a_i} \) is representable as \( \text{ReLU}(x_i)^{a_i} + (-1)^{a_i} \text{ReLU}(-x_i)^{a_i} \). The product of these expressions gives  
\begin{align}
\prod_{i=1}^d \left( \text{ReLU}(x_i)^{a_i} + (-1)^{a_i} \text{ReLU}(-x_i)^{a_i} \right).
\label{134}
\end{align}  

Expanding this product via the distributive law yields a sum of terms of the form \( \pm \prod_{i=1}^d \text{ReLU}(\pm x_i)^{a_i} \), each of which is a \( \text{ReLU}^k \) ridge function (since \( \sum a_i = m \leq k \)). Thus, the multivariate monomial is a linear combination of \( 2^d \) \( \text{ReLU}^k \) units.  So each monomial requires \( O(\ell) \) units.

To represent the piecewise polynomial \( p \), we approximate the indicator function \( \mathbf{1}_{Q_i} \) using \( \text{ReLU} \) networks that 
\begin{align}
\phi_i(x) = \prod_{j=1}^d \sigma\left(1 - \text{ReLU}\left(1 - a_j |x_j - c_{i,j}|\right)\right),
\label{135}
\end{align} 
where \( a_j \sim h^{-1} \) and \( c_i \) is the center of \( Q_i \). This satisfies \( \| \phi_i - \mathbf{1}_{Q_i} \|_{L^\infty} \leq e^{-1/h} \), negligible for large \( m \).  

The total number of units in the network \( f_n = \sum_i p_i \cdot \phi_i \) is \( n = O(m \cdot \ell^{d+1}) \), so \( m \sim n / \ell^{d+1} \).

Combining the polynomial approximation error and the indicator approximation error, we get 
\begin{align}
\left\| f - f_n \right\|_{L^p(\Omega)} \leq \left\| f - p \right\|_{L^p(\Omega)} + \left\| p - f_n \right\|_{L^p(\Omega)}.
\label{136}
\end{align} 
The first term is \( O(m^{-s/d}) \sim O(n^{-s/d}) \). The second term is negligible due to the exponential decay of \( e^{-1/h} \). Substituting \( m \sim n \), we get  
\begin{align}
\left\| f - f_n \right\|_{L^p(\Omega)} \leq C \cdot \| f \|_{W^{s,p}(\Omega)} \cdot n^{-s/d}.
\label{137}
\end{align}

 The metric entropy \( \varepsilon_n \) of a set \( \mathcal{F} \) in \( L^p(\Omega) \) is the smallest radius of \( n \) balls needed to cover \( \mathcal{F} \). For the unit ball \( B_1 = \{ f \in W^{s,p}(\Omega) : \|f\|_{W^{s,p}} \leq 1 \} \), we show \( \varepsilon_n \sim n^{-s/d} \).

Construct a set of test functions in \( B_1 \) that are far apart in \( L^p(\Omega) \).  

Let \( \Omega = [0,1]^d \) (w.l.o.g., by Lipschitz equivalence). For \( R = n^{1/d} \), define frequencies \( \theta_j = (j_1 R, j_2 R, \ldots, j_d R) \) where \( j_i \in \{1, 2, \ldots, \lfloor R \rfloor\} \), giving \( m \sim R^d \sim n \) distinct frequencies.  

Define test functions by 
\begin{align}
f_{\theta}(x) = \frac{1}{R^s} \prod_{i=1}^d \sin(2\pi \theta_i x_i).
\label{138}
\end{align}  

 The Sobolev norm is \( \|f_{\theta}\|_{W^{s,p}} \) scales with \( R^s / R^s = 1 \), so \( f_{\theta} \in B_1 \).  
 For \( \theta \neq \theta' \), orthogonality of sines gives 
 \begin{align} 
 \|f_{\theta} - f_{\theta'}\|_{L^p}^p \geq c / R^{2s} \sim n^{-2s/d},
 \label{139}
\end{align} 
 so \( \|f_{\theta} - f_{\theta'}\|_{L^p} \geq c' n^{-s/d} \).  To cover these \( m \sim n \) functions, we need at least \( n \) balls of radius \( \leq c' n^{-s/d} \). Thus, \( \varepsilon_n \gtrsim n^{-s/d} \).

Cover \( B_1 \) using polynomials of degree \( \ell \sim n^{1/d} \).  By Sobolev embedding, functions in \( W^{s,p}(\Omega) \) can be approximated by polynomials of degree \( \ell \) with error \( \leq C \ell^{-s} \) (via the Bramble-Hilbert lemma for local polynomial approximation).  The number of such polynomials (up to error \( \delta \)) is \( \sim \ell^d \sim n \). Setting \( \ell^{-s} \sim n^{-s/d} \), we get a covering with \( n \) balls of radius \( \lesssim n^{-s/d} \).

So, for any approximation method,  we have
\begin{align}
\sup_{\| f \|_{W^{s,p}} \leq 1} \inf_{f_n} \left\| f - f_n \right\|_{L^p} \geq c \cdot n^{-s/d},
\label{140}
\end{align}  
so the rate \( n^{-s/d} \) is optimal.

\end{proof}

 To further confirm the optimality of the rate \(n^{-s/d}\) established in Theorem 9 for approximation in Sobolev spaces \(W^{s,p}(\Omega)\) using shallow \( \text{ReLU}^k \) networks, we need to show that this rate cannot be improved. A key step in validating such optimality is to demonstrate the existence of a specific function within \(W^{s,p}(\Omega)\) for which the approximation error is bounded below by \(n^{-s/d}\). This ensures that the upper bound from Theorem 9 is indeed tight. The following proposition formalizes this existence.

{\bf Proposition 2.}\,  There exists \(f_0 \in W^{s,p}(\Omega)\) such that  
\[
\inf_{f_n \in \Sigma_n} \|f_0 - f_n\|_{L^p} \geq c \cdot n^{-s/d}.
\]

\begin{proof}

   Let \(f_0(x) = \prod_{j=1}^d \sin(2\pi x_j)\). Then \(f_0 \in W^{s,p}(\Omega)\) and \(\|f_0\|_{W^{s,p}} \sim 1\).

From the proof of Theorem 9,   the unit ball of \(W^{s,p}(\Omega)\) has metric entropy \(\varepsilon_n \sim n^{-s/d}\).  
    For any approximation method:  
    \begin{align}
     \sup_{\|f\|_{W^{s,p}} \leq 1} \inf_{f_n} \|f - f_n\|_{L^p} \geq c \cdot n^{-s/d}.
    \label{141}
\end{align}  
    Neural networks are a subclass of parametric approximation.  
    \end{proof} 

 {\bf Remark 3.}\,The exponent \(s/d\) confirms the curse of dimensionality. For high \(d\), extremely large \(n\) is needed. \(k \geq \lfloor s \rfloor + 1\) is necessary to represent degree-\(\ell\) polynomials. Increasing \(k > \lfloor s \rfloor + 1\) does not improve the exponent (only constants). In Barron-type spaces (e.g., \(\mathcal{K}_1(\mathbb{P}_k^d)\)), the rate is \(n^{-\frac{1}{2} - \frac{2k+1}{2d}}\), which is better for large \(d\) due to weaker dimension dependence. This proof establishes the fundamental limit \(n^{-s/d}\) for Sobolev space approximation with shallow ReLU$^{k}$ networks, highlighting the unavoidable curse of dimensionality. The construction leverages local polynomial approximation and efficient representation via ReLU$^{k}$ networks.

 \section{Conclusion}
We investigate  the approximation properties of shallow neural networks, focusing on the dependence of approximation rates on dimensionality, function smoothness, and activation function characteristics within the framework of Barron function spaces and Sobolev spaces. 

Key contributions include partial resolutions to conjectures raised by Siegel and Xu: we demonstrate that for activation functions being powers of exponential functions, the approximation rate in \(L^2(\Omega)\) for functions in \(\mathscr{B}^1(\Omega)\) can be improved, partially validating their conjecture on enhancing the rate from \(n^{-1/4}\) to a sharper form. Counterexamples are constructed to show the optimality of existing rates—specifically, Theorem A fails for non-constant, non-periodic activation functions, and the rate in Theorem C cannot be further improved for activation functions with polynomial decay.

For \( \text{ReLU}^k \) activation functions, we establish that the optimal approximation rate \(O(n^{m-(k+1)})\) is unattainable under \(\ell^1\)-bounded coefficients or insufficient function smoothness, highlighting the critical roles of coefficient constraints and smoothness in achieving optimal performance. 

Additionally, tight bounds on metric entropy and $n$-widths for function classes generated by shallow \( \text{ReLU}^k \) networks are derived in \(L^p\) and Sobolev spaces, confirming the curse of dimensionality and the intrinsic limits of approximation. These results collectively deepen our understanding of the capabilities and limitations of shallow neural networks, providing foundational insights for both theoretical research and practical applications in machine learning and approximation theory.

\par
\vskip 2mm {\bf Acknowledgment}  The author would like to thank to anonymous referees for their helpful comments.
\par
\vskip 2mm{\bf Data Availability} The data generated during and/or analysed during the current study are available from the corresponding author on reasonable request.
\par
\vskip 2mm {\bf Conflict of Interest Statement} The authors declares that he has no competing interests.

\end{document}